\documentclass[11pt]{article}

\usepackage[preprint]{acl}

\usepackage{times}
\usepackage{latexsym}

\usepackage[T1]{fontenc}
\usepackage[utf8]{inputenc}

\usepackage{microtype}

\usepackage{inconsolata}

\usepackage{graphicx}

\usepackage{pdfpages}
\usepackage{pgffor}     % Provides the \for each loopfigures
\usepackage{rotating}
\usepackage{enumitem}
\usepackage{amsmath, amssymb}
\usepackage{amsthm}
\newtheorem{definition}{Definition}
\usepackage{mathrsfs}
\usepackage{subcaption}
\usepackage{xcolor}
\usepackage{tabularx, booktabs}
\usepackage{tcolorbox}
\usepackage{caption}
\usepackage{algorithm}
\usepackage{algpseudocode}
\definecolor{OliveGreen}{rgb}{0,0.6,0}

\definecolor{ckcolor}{rgb}{0,0,0.6}

\definecolor{c_resid}{HTML}{404040} % Dark Gray for Residuals (Base)
\definecolor{c_att}{HTML}{1F77B4}   % Steel Blue for Attention
\definecolor{c_mlp}{HTML}{D62728}   % Brick Red for MLP

\title{Tracing Computation Density in LLMs}

\author{
 \textbf{Corentin Kervadec\textsuperscript{1}},
 \textbf{Iuliia Lysova\textsuperscript{1}},
 \textbf{Iuri Macocco\textsuperscript{1}},
 \textbf{Marco Baroni\textsuperscript{1,2}},
 \textbf{Gemma Boleda\textsuperscript{1,2}}
\\
 \textsuperscript{1}Universitat Pompeu Fabra
 \textsuperscript{2}ICREA
\\
 \small{
   \textbf{Correspondence:} \href{mailto:corentin.kervadec@gmail.com}{corentin.kervadec@gmail.com}
 }
}

\begin{document}

% models
\newcommand{\qwen}[0]{\texttt{Qwen3-8B-Base}}
\newcommand{\olmo}[0]{\texttt{OLMo2-7B}}
% dataset
\newcommand{\wikitext}[0]{\texttt{Wikitext}}
\newcommand{\prompts}[0]{\texttt{Prompts}}

\maketitle
\begin{abstract}
Transformer-based large language models (LLMs) are comprised of billions of parameters arranged in deep and wide computational graphs, but it is not clear that they exploit their full capacity for all inputs. We introduce the \textit{s-Trace} method to efficiently estimate a subgraph of size \textit{s} that approximates a full model output. With this method, we find the computation in a variety of LLMs to be organized in two distinct phases. A small subgraph mostly composed of early-layer nodes can reconstruct the head of the full model output distribution. Adding further nodes, mostly located in later layers and increasingly consisting of attention heads, leads to incremental refinements in approximating the full output distribution. We find moreover that the amount of necessary computation per input correlates with model uncertainty, and that sparser subgraphs encode shallow statistics, such as unigram frequency. Overall, our results suggest a consistent modular organization in effective LLM computation, with a sparse early-layer core providing a rough prediction that is further refined through denser computations in later layers.\footnote{Code and data available here: \url{https://github.com/CorentinKervadec/llm-strace}.}
\end{abstract}

\section{Introduction}

Modern transformer-based large language models (LLMs) can be represented as deep and wide computational graphs whose execution entails billions of operations that collectively update internal representations to yield a prediction. While standard inference executes the full graph for every input, it remains unclear whether the \textit{effective} computation required to generate a faithful prediction utilizes the model's full capacity or relies on a considerably sparser sub-network. Furthermore, the structural nature of this effective computation is largely unknown: is it uniformly distributed across the network, or is it composed of distinct dense and sparse subgraphs? It is also an open question whether this effective amount of computation is dependent on specific input properties.

Answering these questions has both practical and theoretical implications. From a practical point of view, precisely quantifying the true sparsity of LLM computation and predicting which inputs require more computation can help building more efficient models through pruning, routing, distillation or other techniques. Establishing sparsity is also a prerequisite for meaningful application of interpretability techniques such as circuit analysis, that assume that many language/knowledge processes in LLMs can be reduced to the mechanics of a sparse subgraph \citep{cammarata2020thread:,Ferrando:etal:2024,ameisen2025circuit}. From a theoretical point of view, our pursue bears on two fundamental questions in cognitive science. On the one hand, language competence has traditionally been modeled as a sparse, symbolic, rule-based system greatly at odds with LLMs' highly distributed nature \cite{smolensky2022neurocompositional}. If LLMs turned out to rely on much sparser circuits during average processing, this would pave the way for a reconciliation between the symbolic and distributed views of language~\cite{boleda-2025-llms}. On the other hand, the nature of linguistic complexity and how to measure it are fundamental topics in linguistics \citep[e.g.,][]{Jackendoff:Wittenberg:2014}. If the amount of effective computation in LLMs depends on specific properties of their inputs, then LLMs could provide a practical and precise way to measure complexity.

We analyze 10 different LLMs, ranging from 7B to 14B parameters, and find pervasive sparsity in LLM computation.
By analyzing the trace across varying granularities, we identify two distinct phases. First, for highly sparse subgraphs (beginning at $10^{-4}$ of full model size), LLMs undergo a \textit{construction phase}, in which each operation strongly affects the output prediction, and there is a predominance of early layers being added. Within this phase (at $\approx 10^{-3}$), a highly stable \textit{minimal core} emerges, sufficient to accurately recover the head of the output distribution. Beyond $10^{-2}$, having already reconstructed $40\%$ to $60\%$ of the prediction probability mass, the trace enters a \textit{refinement phase}. Here, the marginal utility of additional operations, which are mostly located in later layers, becomes diffuse. At this stage, the network only fine-tunes the prediction, integrating the nuances required to encode the full richness of the final output distribution.

We also find that different LLMs distribute computation in strikingly similar patterns across similar inputs, suggesting that resource allocation depends on intrinsic input properties, more than on the architectural details of specific models. Two results suggest that computation density, as measured by the \textit{s-Trace} method, is indeed related to the complexity of the input: 1) the density distribution is positively correlated with LLM output entropy (LLMs' own uncertainty estimate); 2) density increases for rarer words (plausibly harder to guess).

\section{Background}

\paragraph{Sparsity in LLMs}
Our work builds on a rich literature establishing structural sparsity in neural networks~\cite{lecun1989optimal, Frankle:Carbin:2019}, which extends directly to LLMs~\cite{Voita:etal:2023}. While static pruning techniques identify permanently redundant parameters or layers across a model~\cite{michel2019sixteen,  lad2024remarkable, li2024adaptive}, a growing body of evidence suggests that computation density is highly dynamic and input-dependent. Modern efficiency frameworks exploit this by dynamically bypassing modular units, attention heads, or entire layers on the fly based on token-level complexity and input difficulty~\cite{zhang2022moefication,liu2023deja, fan2024not, lee2025well}. While these methods focus on task-specific efficiency, our goal is analytical: we seek to quantify the computational footprint required by an LLM to faithfully reconstruct its own output distribution during generic next-token prediction.

\paragraph{Mechanistic Interpretability}
To quantify the amount of computation in LLMs, we leverage methods from mechanistic interpretability. This line of research generally views LLMs as computational graphs composed of discrete, human-interpretable ``circuits" \cite{olah2018building, cammarata2020thread:, ameisen2025circuit}. Notable examples include circuits responsible for the greater-than operation~\cite{hanna2023does}, indirect object identification~\cite{wang2022interpretability}, subject-verb agreement~\cite{Ferrando:etal:2024}, and knowledge retrieval~\cite{meng2022locating}. Circuit discovery can be automated, for instance using greedy search \cite{Conmy:etal:2023}, sparse dictionary learning \cite{bricken2023monosemanticity}, activation patching \cite{syed2024attribution}, or gradient-based pruning \cite{bhaskar2024finding}. More recently, the granularity of circuit discovery has transitioned from coarse attention heads to individual features using sparse autoencoders~\cite{bricken2023monosemanticity}  and cross-layer transcoders~\cite{ameisen2025circuit}. We specifically build upon the Information Flow Route (IFR) framework~\cite{ferrando2024information}, which leverages attribution to efficiently identify circuits at scale. However, rather than applying it to the identification of task-specific circuits, we repurpose it as a general method to estimate the maximum degree of output reconstruction that can be achieved by a subgraph of a certain size.

\paragraph{Structure of Computation in LLMs}
Our \textit{s-Trace} approach is motivated by the ``unraveled view'' of residual networks~\cite{veit2016residual}, which states that the transformer architecture behaves not as a monolithic pipeline, but as an implicit ensemble of shallower subnetworks. Our method exploits this by using an unconstrained graph traversal algorithm to identify and extract these non-sequential, unraveled paths without forcing a rigid layer-by-layer order. This aligns with the iterative inference hypothesis~\cite{jastrzebski2018residual, belrose2023eliciting}, which posits that each residual update in the transformer incrementally modifies a token’s hidden state to progressively shape the next-token distribution~\cite{geva2022transformer}. In contrast, \citet{lad2026remarkable} identified hierarchical ``stages of inference'' where processing goes through four sequential, depth-dependent phases: detokenization, feature engineering, prediction ensembling, and residual calibration. Our study helps validate and reconcile both of these views by identifying distinct phases in the trace.

\section{Methodology}

\label{sec:methodology}

\subsection{LLM as a Causal Graph}

Our work is grounded in the formal representation of a language model's internal processes as a computational graph \cite{Conmy:etal:2023,ferrando2024information, syed2024attribution}. We consider an LLM, that accepts $n$ tokens as inputs, ${t} = (t_1, \dots, t_n)$, and computes a probability distribution over the next token, $P(t_{n+1} \mid {t})$. This process is modeled as a forward pass through a directed acyclic graph $\mathcal{G}=(\mathcal{V},\mathcal{E})$; nodes $\mathcal{V}$ represent intermediate states of the tokens and edges $\mathcal{E}$ represent operations affecting the information flow.

We focus on the standard transformer architecture~\cite{vaswani2017attention}. 
The forward pass consists of a sequence of residual blocks (or layers). First, discrete tokens are embedded into continuous vectors, $h^0 = (h^0_1, \dots, h^0_n)$, forming the initial nodes of the graph. These representations are iteratively transformed through $L$ layers. The state at layer $l$, denoted $h^l = (h^l_1, \dots, h^l_n)$, is a function of the previous layer's activations.

Each layer $l$ is itself composed of two sub-layers: a Multi-Head Attention (MHA) and a Multi-Layer Perceptron (MLP). 
For a specific token $i$ at layer $l$, the MHA update rule to produce the sub-layer output ${z}^{l}_i$ is:
\begin{equation}
    \label{eq:graph_view}
    {z}^{l}_i \;\; = \underbrace{\textcolor{c_resid}{h^l_i}}_{\textcolor{c_resid}{\text{Residual Edge}}} \!\!\! + \;\;\;  \sum_{k, j} \underbrace{\textcolor{c_att}{\phi^{l,k}(h^l_i, h^l_j)}}_{\textcolor{c_att}{\text{Attention Edges}}}
\end{equation}
where $k\in[1, N_H]$ with $N_H$ the number of attention heads, $j\in[1,n]$, and $\phi^{l,k}(h^l_i, h^l_j)$ representing an \textit{attention head contribution}: the vector ``moved'' from source token $j$ to target token $i$ by head $k$ \citep{ferrando2024information}. This pairwise decomposition is central to our graph formulation. Similarly, the MLP update rule is:
\begin{equation}
    \label{eq:graph_view_mlp}
    h^{l+1}_i \;\; = \underbrace{\textcolor{c_resid}{{z}^{l}_i}}_{\textcolor{c_resid}{\text{Residual Edge}}} \!\!\! + \;\;\; \underbrace{\textcolor{c_mlp}{\text{mlp}_l(z^{l}_i)}}_{\textcolor{c_mlp}{\text{MLP Edge}}}
\end{equation}

By unrolling Equations \ref{eq:graph_view} and \ref{eq:graph_view_mlp}, we observe that the vector representation at any node is simply the sum of vectors from its incoming edges. This yields a granular view of information flow. An MHA sub-layer is decomposed into $N_H \times n + 1$ edges per target token,\footnote{In causal masking, some edges are set to 0.} and an MLP sub-layer gives rise to $2$ edges per target token.

Formally, the computational graph $\mathcal{G}$ is defined as in Table~\ref{tab:graph_def}. Note that both nodes \textit{and} edges are defined token-wise. This allows for finer modeling of information routing compared to approaches that aggregate head outputs into a single edge per target token \citep[e.g.,][]{syed2024attribution}. As an illustration, given an input of length $100$, a model with 32 attention heads and 32 layers corresponds to a graph with $6,302$ nodes and $5,022,103$ edges. 

\begin{table}[t]
    \centering
    \small
    \begin{tabularx}{\columnwidth}{l X}
    \toprule
    \textbf{Component} & \textbf{Formal Definition of $\mathcal{G}$} \\
    \midrule
    Nodes $\mathcal{V}$ & $\{h^l_{i+1}, z^l_{i+1} \mid l \in [0, L], i \in [1, n]\}$ \\
    \addlinespace[0.5em]
    Edges $\mathcal{E}$ & 
        $\begin{cases} 
        \text{Attn:} & \phi^{l,k}(h^l_i, h^l_j) \text{\hspace{0.6cm} (Eq.~\ref{eq:graph_view})}\\
        \text{MLP:} & \text{mlp}_l(z^l_i)
        \text{\hspace{0.99cm} \,(Eq.~\ref{eq:graph_view_mlp})}\\
        \text{Resid:} & h^l_i, z^l_i
        \end{cases}$ \\
    \bottomrule
    \end{tabularx}    
    \caption{Components of the computational graph $\mathcal{G}$.}
    \label{tab:graph_def}
\end{table}

\subsection{s-Trace Definition}

\begin{definition}[$s$-Trace] 
\label{def:trace}
Given a budget $s \in \mathbb{N}^+$, the \textit{s-Trace} $\mathcal{T}_s = (\mathcal{V}_s, \mathcal{E}_s)$ is an approximation of the optimal subgraph of $\mathcal{G}$ such that $\mathcal{V}_s \subseteq \mathcal{V}$, $\mathcal{E}_s \subseteq \mathcal{E}$, $|\mathcal{E}_s| = s$, and:
\begin{equation*}
    \mathcal{E}_s = 
    \operatorname*{arg\,min}_{\mathcal{E}_s \subset \mathcal{E}, \; |\mathcal{E}_s|=s}
    \mathcal{D}\left( P(t_{n+1} \mid {t}), \hat{P}_{\mathcal{T}_s}(t_{n+1} \mid {t}) \right)
\end{equation*}
where $\mathcal{D}$ is a distance metric and $\hat{P}_{\mathcal{T}_s}$ is the prediction restricted to subgraph $\mathcal{T}_s$.
\end{definition}

In words, the trace  $\mathcal{T}_s$ approximates the subgraph of size $s$ whose predictions are closest, among all possible subgraphs of the same size, to those of the full model.
During inference, the s-Trace's prediction $\hat{P}_{\mathcal{T}_s}$ is obtained by applying a binary mask, $M = \{m_e\}_{e \in \mathcal{E}}$, to the computational graph, where $m_e = 1$ if edge $e \in \mathcal{E}_s$ and $0$ otherwise.
For instance, the masked update rule for the MHA sub-layer (Eq.~\ref{eq:graph_view}) becomes:
\begin{equation}
    \label{eq:att_ablation}
    {z}^{l}_i = m^{l}_{i, \text{res}} \cdot \textcolor{c_resid}{h^l_i} + \sum_{k, j} m^{l,k}_{i,j} \cdot \textcolor{c_att}{\phi^{l,k}(h^l_i, h^l_j)}
\end{equation}
where $m \in \{0, 1\}$ are the binary decisions for the residual and attention edges respectively. 
In the rest of the paper, we abuse the size symbol $s$ to refer to the relative number of edges $\bar{s} = |\mathcal{E}_s|/|\mathcal{E}| \in [0,1]$. This quantity more intuitively denotes the size of the chosen subgraph relative to the full model.

\subsection{Trace Extraction via Graph Traversal}
\label{sec:extraction}
The challenge of trace extraction is to identify the subset of edges, $\mathcal{E}_s \subset \mathcal{E}$ with $|\mathcal{E}_s|=s$, that best reconstruct the model's output distribution (cf. Def.~\ref{def:trace}). 
Because of the large scale of the models analyzed (7B to 14B parameters) and the dense and intricate nature of the computational graph, identifying the \textit{optimal} trace is intractable. Given that computationally expensive methods like greedy search~\cite{Conmy:etal:2023} or activation patching~\cite{syed2024attribution} are infeasible for identifying all edges, we approximate the trace with a training-free, activation-based strategy that just requires a single pass. 

\paragraph{Importance Score}
We compute the importance score associated to each edge $e \in \mathcal{E}$ using the L1-norm, positing that the impact of an operation is proportional to the magnitude of its update vector. To ensure fair comparison across the graph, scores are normalized node-wise.
For any edge $e$ carrying a vector $\mathbf{v}_e$ (e.g., an attention head output $\phi^{l,k}$) into a target node $v$, the score is computed relative to all other edges feeding into $v$, such that $\mathcal{I}(e) = \| \mathbf{v}_e \|_1 / \sum_{e' \rightarrow v} \| \mathbf{v}_{e'} \|_1$. We experimented with alternatives, such as L2-norm or similarity between the node and the edges, and found the L1-norm to be either comparable or superior (yielding lower error at same sizes, cf. Fig~\ref{fig:alternative_scores}).

 \paragraph{Graph Traversal}
Trace extraction is based on \textit{Greedy Best-First Search}, a graph traversal method that iteratively increases the trace size by adding the most important edge available, starting from the last-token node in the last layer, until the budget $s$ is reached.
Every time an edge $e$ is added, the nodes visited by $e$ are added, too, and the edges connected to these nodes become available for selection. App.~\ref{app:graph-algo} provides the full definition and Fig.~\ref{fig:graph_construction_part1}-\ref{fig:graph_construction_part2} an illustration.

\begin{figure*}[p] % [p] tells LaTeX to try putting this on a dedicated float page. Use [t] if you want text below it.

    % --- THE TABLE ---
    \centering
    \small
    \begin{tabularx}{\textwidth}{l X r}
    \toprule
    \textbf{Phase} & \textbf{Description} & \textbf{Scale} \\
    \midrule
    \textbf{Construction Phase} & Steep decay in reconstruction error, marking the initial assemblage of the ``core'' architecture. Driven by a sharp increase of early-layer paths. & $s\in[10^{-4}, 10^{-2}]$ \\
    \addlinespace[0.6em]
    \textbf{Minimal Core} & A highly stable, architecture-invariant early-layer graph backbone that is causally sufficient for recovering the salient modes of the prediction. & $s\approx 10^{-3}$ \\
    \addlinespace[0.6em]
    \textbf{Refinement Phase} & Progressively incorporates the nuanced information required for recovering the full prediction. Emergence of distributed, late-layer attention paths. & $s> 10^{-2}$ \\
    \bottomrule
    \end{tabularx}
    \captionof{table}{Important stages in trace size profiles. See formal definition in Appendix~\ref{app:formal_phases}.}
    \label{tab:trace_taxonomy}

    \vspace{15pt} % Breathing room
    
    % --- FIRST FIGURE ---
    \centering
    \includegraphics[width=\linewidth]{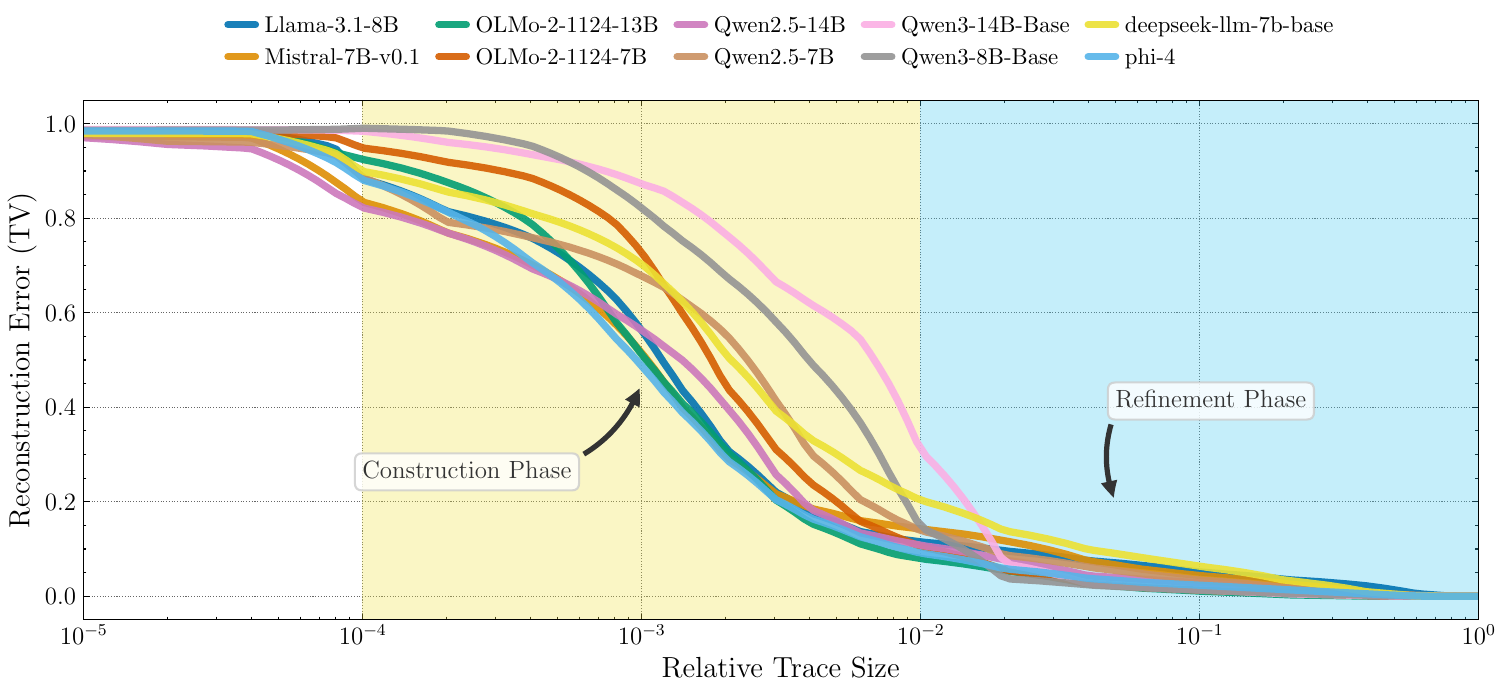}
    \captionof{figure}{Relation between reconstruction error (measured with Total Variation distance) and trace size for various LLMs. The trace undergoes two distinct phases: a \textit{construction phase}, and a \textit{refinement phase}.
    }
    \label{fig:tv_size}
    
    \vspace{15pt} % Add some breathing room between the items
    
    % --- SECOND FIGURE ---
    \centering
    \includegraphics[width=\linewidth]{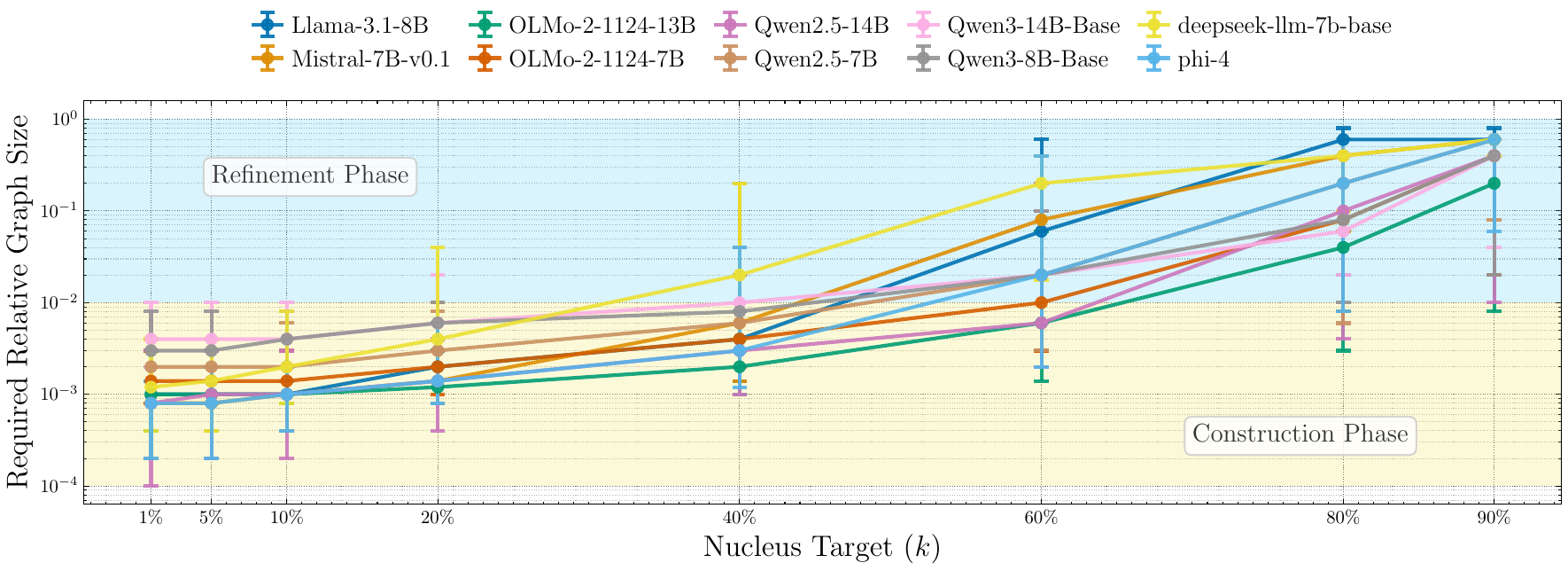}
    \captionof{figure}{The minimal size $s$ required to reconstruct the top-$k\%$ (\textit{nucleus target}) of the full model's probability mass. A very small trace ($s \approx 10^{-3}$, the \textit{minimal core}) suffices to reconstruct the top 1\% prediction, but only at the end of the \textit{construction phase} ($s=10^{-2}$) do we recover $40\%$ to $60\%$ of the probability mass.}
    \label{fig:nu_size}
    
\end{figure*}

\subsection{Evaluating Trace Faithfulness}

To evaluate the quality of a trace, we measure its \textit{reconstruction error}: how well the masked model output $\hat{P}_{\mathcal{T}_s}(t_{n+1} \mid {t})$ approximates the full model output $P(t_{n+1} \mid {t})$. Unlike prior work, that often focuses on binary task success, e.g., whether the correct token is the top-1 prediction~\cite{heimersheim2024use}, we adopt a more general approach by measuring the preservation of the full output distribution. We quantify this using the \textit{Total Variation} (TV) distance:
\begin{equation}
    \label{eq:deltaTV}
    \delta_{TV}(s) = \frac{1}{2} \sum_{v \in \mathbb{V}} |P(v) - \hat{P}_{\mathcal{T}_s}(v)|
\end{equation}
where $\mathbb{V}$ is the token vocabulary, and $\hat{P}_{\mathcal{T}_s}(v)$ and $P(v)$ are probabilities assigned to token $v$. A low $\delta_{TV}$ indicates that the trace is \textit{sufficient} to recover the model's behavior.\footnote{To assess necessity--whether the edges in $\mathcal{T}_s$ are the \textit{only ones} capable of producing the output--we employ a complementary evaluation in App.~\ref{app:suf_nec}, where we ablate the trace edges and observe the corresponding degradation.}
We select TV over alternatives such as the Kullback-Leibler divergence because the latter is unbounded and sensitive to relative errors, meaning that it can be dominated by discrepancies in the low-probability tail.
In contrast, TV measures the absolute difference in probability mass. This ensures that our metric focuses on the dominant modes of the distribution—where the model's actual predictions lie—rather than being disproportionately affected by the noisy tail. 

\subsection{Experimental Details}

\paragraph{Dataset}
We use a dataset of $5,000$ sequences from Wikitext \citep{Merity:etal:2016}. They  were selected to contain 40 words on average (one or two sentences), making sure that they start at a sentence beginning and end with either a full word or punctuation (no word fragment). Following trace extraction, we discard any instances that encountered runtime anomalies or numerical instability (such as out-of-memory errors or NaN values). We get minimally 4,250 successfully processed sequences per model (Table~\ref{tab:model_summary} in App.~\ref{app:model-zoo}).

\paragraph{Model Zoo}
We use 10 different LLMs from 7 HuggingFace families (Table~\ref{tab:model_summary} in App.~\ref{app:model-zoo}), with sizes ranging in $[7, 14]$ billion parameters.

\paragraph{Trace Collection}
For each input, we extract the s-Trace at 26 levels of granularity, from very small to very large, through a fixed size grid (App.~\ref{app:grid-size}).
We collect about $1.3$M traces in total.

\paragraph{Baseline}
\label{ssec:baseline}
We compare the s-Trace against a strong random baseline. Rather than performing a naive uniform random pick, we made sure to always preserve residual connections, and MLPs are explicitly prioritized over attention edges. Because MLP edges are heavily outnumbered by attention edges, a completely uniform random baseline would rarely select an MLP at sparse budgets.

\section{Results}

\subsection{Stages of Reconstruction in the Trace}
\label{sec:computation_density}

We first measure the extent to which the trace size influences its reconstruction error. 
If LLM computation is sparse, then a small trace should be able to faithfully reconstruct the full model distribution. A dense computation would result in a larger trace. Our analysis reveals that the reconstruction process follows three distinct steps: a \textit{construction phase}, during which a \textit{minimal core} is established, and a progressive \textit{refinement phase} (Table~\ref{tab:trace_taxonomy}).

\paragraph{Construction Phase} Very small traces ($s < 10^{-4}$) exhibit high reconstruction error (Fig.~\ref{fig:tv_size}). A distinct \textit{construction phase} occurs when $s \in [10^{-4}, 10^{-2}]$ ($0.01\%$–$1\%$ of the graph), characterized by a sharp, non-linear error decay with high marginal edge utility. This suggests that the dense model's functional output is governed by a sparse subgraph present at this threshold. 

\paragraph{Minimal Core}
To better understand the effect of the construction phase on model prediction, we define a \textit{nucleus reconstruction} metric, measuring the minimal size $s$ required to reconstruct the top-$k\%$ of the full model's probability mass, for $k \in \{1, 5, 10, \dots, 90\}$. This metric, inspired by nucleus sampling~\cite{Holtzman2020The}, identifies the computational budget necessary to recover varying amounts of the original distribution. Fig.~\ref{fig:nu_size} shows that a remarkably small subgraph ($s \approx 10^{-3}$) forms a \textit{minimal core} sufficient to recover the top-1\% of the distribution, which includes the most likely token. This subgraph size corresponds to the middle value in the construction phase observed in Fig.~\ref{fig:tv_size}.
While it suffices for greedy decoding, it still lacks the nuance needed for high-quality generation. The latter is reached at the end of the construction phase ($s \approx 10^{-2}$), where the trace is able to reconstruct the top $40\%$ to $60\%$ of the prediction probability mass.\footnote{The typical value for good-quality generation with nucleus sampling is $k=60\%$~\cite{Holtzman2020The}.}

\paragraph{Refinement Phase}
Beyond $s > 10^{-2}$, we observe a ceiling effect where the already low reconstruction error slowly decreases (Fig.~\ref{fig:tv_size}). The model enters a \textit{refinement phase} in which achieving higher fidelity (e.g., $k > 60\%$) requires an order-of-magnitude increase in trace size (Fig.~\ref{fig:nu_size}). In this stage, the marginal utility of adding edges is small, as the trace progressively incorporates the nuanced information required to reconstruct the full distribution. Fig.~\ref{fig:nu_size} shows that, once the top-10\% of the probability mass is reconstructed, $\log(s)$ scales linearly with $k$. This log-linear behavior suggests that, in this regime, the remaining distribution is constructed via the additive accumulation of relatively independent computational paths, rather than a single, densely interconnected component. 

\subsection{Structure of the Trace}
\label{sec:trace_structure}

\begin{figure}[t]
    \centering
    \includegraphics[width=0.9\linewidth, page=7]{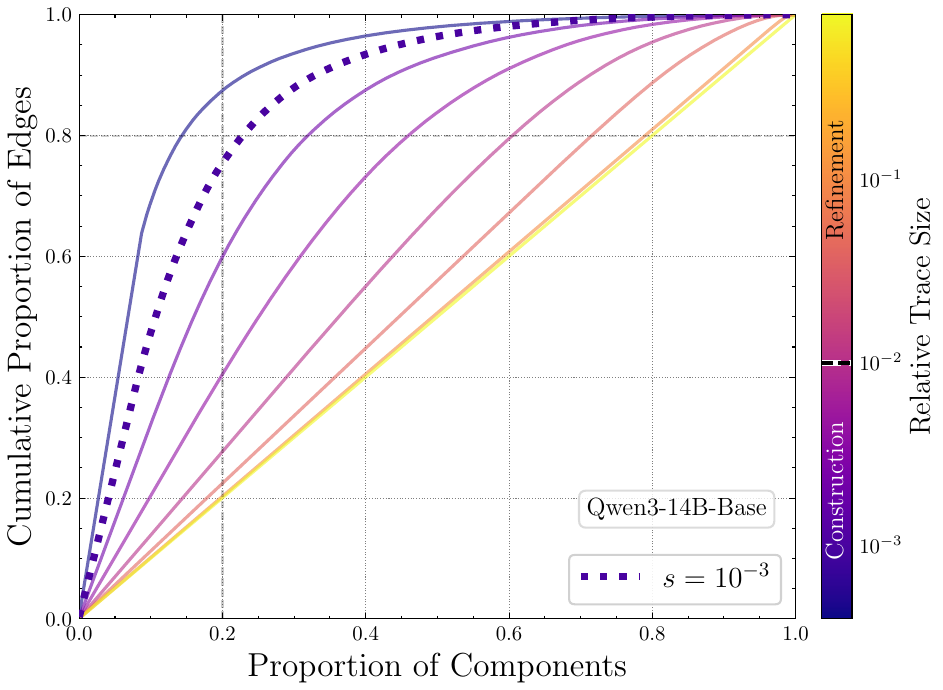}
    \caption{Cumulative edge allocation across components ranked by frequency in the dataset for \texttt{Qwen3-14B-Base} (more LLMs in App.~\ref{app:comp_freq}). At the minimal core ($s = 10^{-3}$, dashed line), edges are concentrated within a small subset of frequent components.} 
    \label{fig:component_freq}
\end{figure}

\begin{figure*}[t]
    \centering
    
    % --- Subfigure A: Layer Composition ---
    \begin{subfigure}{\linewidth}
        \centering
        \caption{\label{fig:layer_size}} % Moved to the top
        \includegraphics[width=0.9\linewidth, page=7]{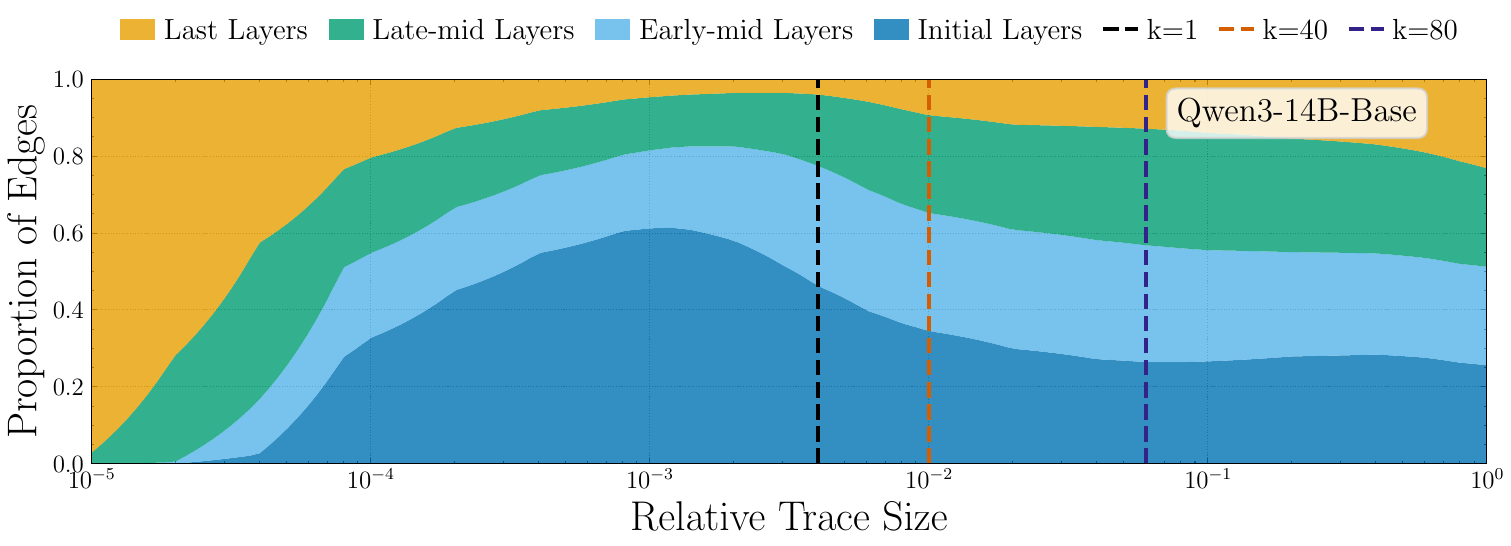}
    \end{subfigure}
    
    \vspace{5pt} % Reduced spacing between plots
    
    % --- Subfigure B: Component Composition ---
    \begin{subfigure}{\linewidth}
        \centering
        \caption{\label{fig:trace_composition}} % Moved to the top
        \includegraphics[width=0.9\linewidth, page=7]{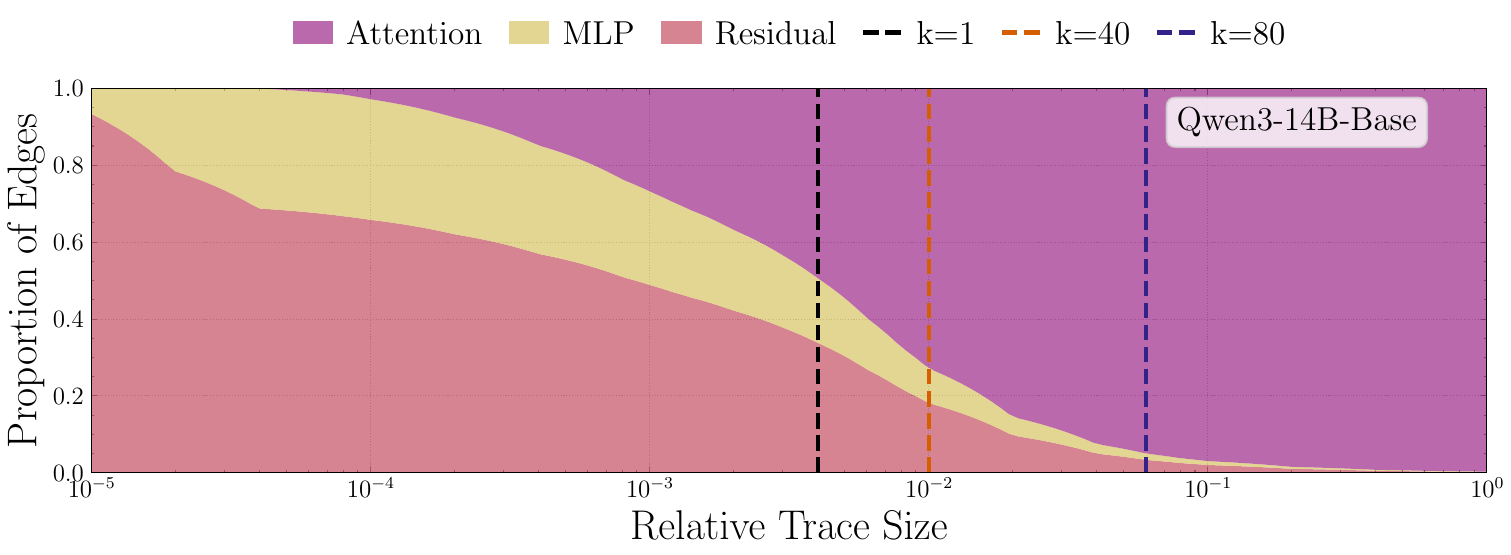}
    \end{subfigure}
    
    % --- Unified Shared Caption ---
    \caption{Trace makeup at different sizes for \texttt{Qwen3-14B-Base} (more in App.~\ref{app:structure}). \textbf{(a)} Composition by layer depth. \textbf{(b)} Composition by transformer components. The vertical lines mark the median size for nucleus $k$ reconstruction.}
    \label{fig:combined_trace_makeup}
    \vspace{-10pt}
\end{figure*}

\paragraph{Stability of the Minimal Core} We ask whether the minimal core is shared across inputs, or whether it is input-dependent.
For each trace size, we calculate the global frequency of each trace component (e.g., residual at layer $l$, MLP at layer $l$, attention head $k$ at layer $l$) across all instances of the dataset, rank them according to frequency, and plot the cumulative edge allocation $y$ accounted for by the top-$x$ fraction of the ranked components (Fig.~\ref{fig:component_freq}).
Around the minimal core, at $s = 10^{-3}$, 
the distribution is highly concentrated: For instance, in \texttt{Qwen1-14B-Base}, the top $20\%$ most frequent components account for $80\%$ of all edges across all dataset instances. This shows the minimal core is remarkably stable across contexts. As the trace size increases, the distribution predictably flattens. 
To verify whether this global concentration holds at the level of individual instances, we evaluate the pairwise component overlap between random pairs of dataset instances. To eliminate artifacts arising from the bounded total number of components in the model, we also measure the statistical significance of the intersection against a null model, reporting the $Z$-score. At the minimal core, we obtain a Jaccard overlap of $\approx0.5$ as well as a $Z$-score of $\approx20$, indicating that the overlap is highly significant.

\paragraph{Layer Makeup at Different Stages} Fig.~\ref{fig:layer_size} illustrates the layer-wise distribution of the transformations that comprise the trace. At a very small size ($s < 10^{-4}$), the trace is dominated almost exclusively by the final layers of the model. However, coinciding with the onset of the construction phase ($s = 10^{-4}$), the proportion of early layers increases sharply, peaking at $s = 10^{-3}$. At this stage, initial layers and early-middle layers together constitute $80\%$ of the edges. This peak closely aligns with the budget required to reconstruct the top-1\% of the probability mass. Beyond this threshold, the proportion of late and late-middle layers steadily increases. Thus, (1)~the minimal core relies predominantly on early-layer representations, and (2)~the subsequent refinement phase ($s > 10^{-2}$) is driven by later layers, that might nuance and contextualize the predictions established earlier.

\paragraph{Extreme Sparsity in Attention}
Fig.~\ref{fig:trace_composition} displays the trace composition categorized into attention, MLP, and residual connections. By design (Sec.~\ref{sec:methodology}), the computational graph is heavily skewed; attention heads account for over $99\%$ of all possible edges (see the far right of the figure, where the distribution for the full model is shown). 
Strikingly, however, 
the minimal core has a remarkably balanced proportion of trace components, with approximately $30\%$ residual edges, $20\%$ MLP edges, and only $50\%$ attention edges. This disproportion implies that attention operates under extreme sparsity, as only a minimal fraction of total attention edges is required. The subsequent refinement phase ($s > 10^{-2}$) is however characterized by an increased contribution of attention, suggesting that the latter plays a key role in refinement.

\begin{figure*}[ht]
    \centering
    \includegraphics[width=\linewidth]{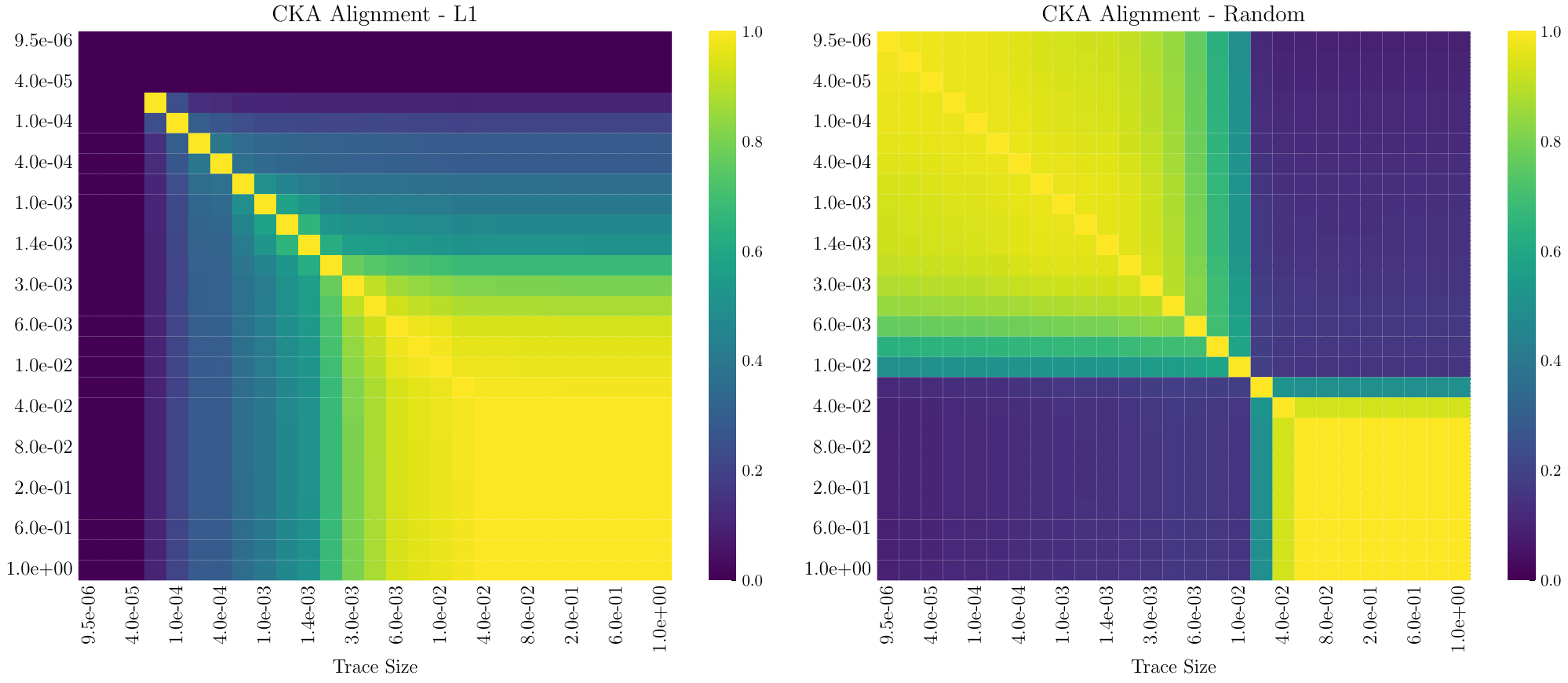}
    \caption{Geometric alignment (CKA) of hidden representations at various trace sizes ($x$ and $y$ axes) against the full model (\olmo). A score close to $1.0$ (\textit{yellow}) indicate highly aligned representations. Our $s$-Trace method (based on the $L1$ importance) is shown in the left panel. It is compared with the baseline (see Section~\ref{ssec:baseline}) shown in the right panel. We observe that $s$-Trace captures a gradual, meaningful characterization of the internal representations.}
    \label{fig:cka}
\end{figure*}

\subsection{Input-related Effects on Sparsity}
\label{sec:sparse_context}

We next investigate which contextual factors modulate computation density. We define a quantity $\mathcal{C}$, obtained by measuring, for each input separately, the area under the curve (AUC) of the error-vs-trace-size plot (App.~\ref{app:rho}). The larger this area, the more resources an LLM is effectively recruiting to produce its output given the current input.

\paragraph{Consistency Across LLMs} 
The observed computation density patterns across inputs are remarkably stable across LLMs, with positive correlations ranging from $0.26$ to $0.71$ ($\mu=0.52$, $sd=0.11$, full results in App.~\ref{app:pairwise_corr}). This strong alignment indicates that computation density is largely independent of architecture or parameter scale. 
% note from G: I extracted the cor values from the figure in the appendix
% and computed sd with R:
% cor_values <- c(
%  0.69,
%  0.54, 0.44,
%  0.53, 0.46, 0.63,
%  0.62, 0.56, 0.58, 0.55,
%  0.64, 0.58, 0.60, 0.57, 0.71,
%  0.40, 0.29, 0.50, 0.48, 0.47, 0.53,
%  0.43, 0.36, 0.50, 0.50, 0.44, 0.54, 0.60,
%  0.61, 0.67, 0.38, 0.39, 0.48, 0.53, 0.26, 0.32,
%  0.66, 0.58, 0.51, 0.51, 0.63, 0.59, 0.39, 0.41, 0.53
%)
% mean(cor_values) # 0.5153333
% sd(cor_values) # 0.1060789

\paragraph{More Uncertainty, More Computation}
We quantify prediction uncertainty using the Shannon entropy of the output probability distribution of the full model and correlate it with the amount of computation $\mathcal{C}$. Across all models, this correlation is significantly positive, albeit moderate, spanning $[0.04, 0.37]$ ($\mu=0.22$; full results in App.~\ref{app:entropy}). For low-entropy contexts, minimal traces achieve near-perfect reconstruction, indicating high sparsity. Conversely, high-entropy contexts require larger trace budgets to achieve equivalent error, characterized by an extended refinement phase.

\paragraph{Sparser Traces Encode Shallow Statistics} 
We further investigate how the (Wikitext-derived) frequency of the highest-probability output token  relates to computation density. We observe a consistently negative correlation between token frequency and $\mathcal{C}$ across all evaluated model families, with correlations spanning $[-0.45, -0.22]$ (full results in App.~\ref{app:entropy}). This indicates that more frequent tokens tend to be associated with less computation, and that the relationship is robust across different models. This suggests that sparse traces are sufficient to encode superficial statistics, such as unigram frequency. Capturing more complex linguistic dependencies, such as those required to predict rare tokens, necessitates larger traces.

\subsection{How good is magnitude-based trace extraction?}
\label{sec:magnitude_trace_evaluation}

As noted in Section~\ref{sec:methodology}, $L_1$-based trace extraction is not strictly optimal. Rather, it is a scalable empirical proxy that provides an upper bound on how small a faithful subgraph can be.

\paragraph{Why not use causal patching?}
Causal patching \citep[e.g.,][]{syed2024attribution} relies on hand-crafted contrastive pairs to isolate task-specific circuit mechanisms. In contrast, our goal is to analyze the continuous size-vs-fidelity curve of general language modeling, for which a universal baseline for counterfactual states does not exist.
Moreover, patching aims to uncover highly localized circuits. Conversely, our study requires tracking subgraphs spanning five orders of magnitude ($0.001\%$ to $100\%$ of full model size). Running an edge-patching protocol over millions of edges on a 14B parameter model across thousands of tokens is computationally intractable.
Admittedly, magnitude heuristics lack direct causal guarantees. However, as recently highlighted by \citet{meloux2025everything}, many AI interpretability studies lack causal rigor, including patching-based circuit discovery works. We thus intentionally shifted the paradigm: instead of building localized circuits for a niche task, we scaled existing mechanistic interpretability tools to provide general density profiles on a large scale across multiple LLM families. We achieve greater robustness and coverage by trading expensive local causal heuristics for scalable attribution.

\paragraph{Importance score}
We compare the $L_1$ importance score against three alternative scoring functions: $L_2$-norm, proximity score \citep[adapted from][]{ferrando2024information}, and cosine similarity (details are provided in Appendix~\ref{app:alternative_scores}). Magnitude-based metrics consistently outperform alignment-based metrics (cf.\ Fig.~\ref{fig:alternative_scores}).

\paragraph{Geometry of the traces' representation space}
To verify that traces extracted via the $L_1$ heuristic capture intrinsic model properties, we analyzed the geometry of the internal representations across trace sizes. Using CKA across 1,000 instances (on \olmo), we measured the geometric alignment of hidden representations at various trace sizes against the full model (Fig~\ref{fig:cka}). The $L_1$ trace shows a smooth, continuous convergence toward the full model between $s=10^{-4}$ and $10^{-2}$ before stabilizing. Conversely, the baseline (see Section~\ref{ssec:baseline})  fractures into two disconnected blocks with a sharp transition at $10^{-2}$. This confirms that our trace approximation captures a gradual, meaningful characterization of the LLM’s internal representations.

\section{Discussion}

\paragraph{Key Findings}
We find two distinct regimes in how computation is allocated across recent LLMs.
First, spanning from 0.01\% to 1\% of the full model size, the trace exhibits a \textit{construction phase} in which each added operation exerts a major impact on the reconstruction of the prediction. Second, from 1\% to 100\% of the model size, the trace enters a \textit{refinement phase}, where the prediction is slowly fine-tuned, incorporating the subtle nuances required to encode the full richness of the model's output distribution. The construction phase is marked by the emergence of the \textit{minimal core}, which we estimate to be contained within only 0.1\% of the full model. This remarkably compact set of operations appears sufficient for predicting the highest-probability token of the full distribution, while remaining stable across diverse inputs.

Although our \textit{s-Trace} method is depth-agnostic \cite[due to its ``unraveled view'';][]{veit2016residual}, our findings align with a staged view of transformer depth~\cite{lad2024remarkable}. The minimal core is predominantly composed of early-layer operations where detokenization and feature engineering might occur. Conversely, the refinement phase relies increasingly on later layers for prediction ensembling and residual calibration to fine-tune the final output distribution. \citet{csordas2026language} observed that skipping the second half of layers minimally affects representations for \textit{future} token predictions (beyond the next token), indicating that late operations are largely decoupled and focus primarily on polishing the immediate distribution.

\paragraph{Implications for Interpretability}
Our findings offer several takeaways for NLP interpretability:

\textit{(1) What is the appropriate scale for circuits?}
There is a threshold below which circuits fail to faithfully represent the model's internal processing. As a rule of thumb, we suggest considering the minimal core (0.1\%) as the smallest scale for basic faithfulness. While more sophisticated circuit discovery methods might push lower, the minimal core ensures MLP and attention edges are well-balanced, making it easier to isolate their respective roles (Fig.~\ref{fig:trace_composition}).

\textit{(2) Output distributions matter.}
We observe that output distribution entropy closely relates to the model's computational volume. Evaluating subgraphs using simplified task metrics that ignore distribution properties can yield degenerate circuits. 

\textit{(3) Not all predictions are created equal.}
Just like low- and high-entropy predictions elicit different computational profiles, token frequency correlates with effective graph size. This suggests that baseline statistical properties of token prediction must be considered when deriving and analyzing circuits.

\textit{(4) Mechanistic interpretability need not restrict itself to hyper-sparse circuits.} Integrating computational graph formalisms with graph theory tools offers a promising framework for analyzing dense, distributed processing. Our computation density metric represents an initial step. We hope this work encourages the community to explore other network properties, such as centrality or modularity, to better characterize the complex internal dynamics of LLMs beyond simple circuit identification.

\paragraph{Implications for Cognitive Science}
The robust link between computation density and input properties, and the fact that density varies in similar ways across models open exciting avenues for research.
Computational density could potentially serve as a novel metric for \textit{linguistic complexity}. Moreover, our results suggest that LLMs recruit fewer resources for less demanding inputs (such as frequent words), pointing to the idea that LLMs implement a form of \textit{efficient processing}, in line with the emerging consensus on how humans process language~\citep{Gibson:etal:2019}. Last, analyzing which linguistic features modulate computation density can shed light on whether LLMs naturally find a trade-off between more distributed and more symbolic processing, akin to the one conjectured for human language \cite{boleda-2025-llms}.

\newpage

\section*{Limitations}

\textit{(a)} Our findings directly depend on how we identify the trace. Better methods could lead to smaller reconstruction error for equivalent budgets, or highlight emerging patterns we could not identify with the current approach. As a sanity check, we experimentally compared our approach with a strong baseline consisting in a random graph traversal that favors residual and MLP edges, and found that our method provides better results (App.~\ref{app:rand_baseline}).

\textit{(b)} While we evaluate reconstruction error via zero-ablation of the edges, recent work notes that zero-ablation can push activations off the data manifold, a limitation often mitigated by mean-ablation or activation patching \citep{li2024optimal}. As a sanity check, we evaluated zeroing attention edges \textit{before} the softmax (a softer ablation since the softmax operation rebalances weights across remaining edges) and observed results consistent with our primary approach. We ultimately retain the after-softmax ablation for its superior interpretability. We leave more nuanced interventions, such as mean-edge replacement, to future work due to the technical challenges they introduce in our fine-grained setup.

\textit{(c)} Our study is limited to English data and the transformer architecture. We plan in the future to apply it to other languages and architectures to broaden the generality of our results.

\section*{Impact Statement}

This work advances the field of Mechanistic Interpretability, which aims to elucidate the internal decision-making processes of Large Language Models (LLMs). By characterizing how computation is allocated in LLMs, our research contributes to the development of more transparent and accountable AI systems. Understanding the true computation and information flow is a prerequisite for reliable model monitoring, debugging, and safety auditing.
We do not foresee any direct negative societal impacts from this work. While insights into model efficiency could theoretically be used to optimize harmful systems, the primary application of our findings is to prevent oversimplified explanations of model behavior, thereby fostering more robust safety evaluations.

\section*{AI Assistant}
We acknowledge the use of an AI assistant to aid in drafting sections of this manuscript and generating code for the experiments. All AI-generated text and code were thoroughly reviewed, and the final edits and revisions were exclusively performed by the authors.

\section*{Acknowledgment}
We would like to thank our colleagues at COLT for advice
and feedback. CK, IM and MB were funded by the European Research Council (ERC) under the European Union’s Horizon 2020 research and innovation program (grant agreement No. 101019291). This paper reflects the authors’ view only, and the ERC is not responsible for any use that may be made of the information it contains. 

\bibliography{biblio,marco}

%%%%%%%%%%%%%%%%%%%%%%%%%%%%%%%%%%%%%%%%%%%%%%%%%%%%%%%%%%%%%%%%%%%%%%%%%%%%%%%
%%%%%%%%%%%%%%%%%%%%%%%%%%%%%%%%%%%%%%%%%%%%%%%%%%%%%%%%%%%%%%%%%%%%%%%%%%%%%%%
% APPENDIX
%%%%%%%%%%%%%%%%%%%%%%%%%%%%%%%%%%%%%%%%%%%%%%%%%%%%%%%%%%%%%%%%%%%%%%%%%%%%%%%
%%%%%%%%%%%%%%%%%%%%%%%%%%%%%%%%%%%%%%%%%%%%%%%%%%%%%%%%%%%%%%%%%%%%%%%%%%%%%%%

\appendix

\section{Graph Traversal}
\label{app:graph-algo}

\begin{figure*}[t]
    \centering
    % Row 1
    \begin{subfigure}[t]{0.48\textwidth}
        \centering
        \includegraphics[width=\linewidth]{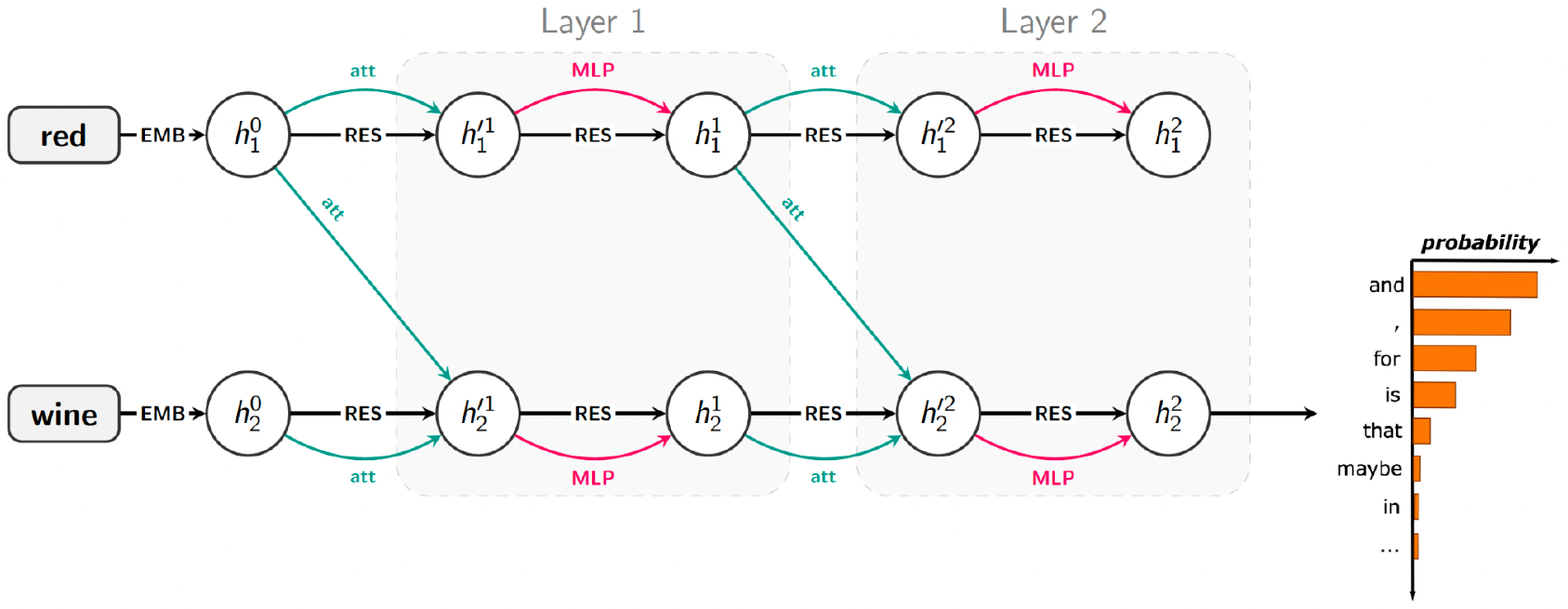}
        \caption{Full model prediction (forward pass). The most important connections (carrying the most weight) are calculated.}
        \label{fig:step1}
    \end{subfigure}\hfill
    \begin{subfigure}[t]{0.48\textwidth}
        \centering
        \includegraphics[width=\linewidth]{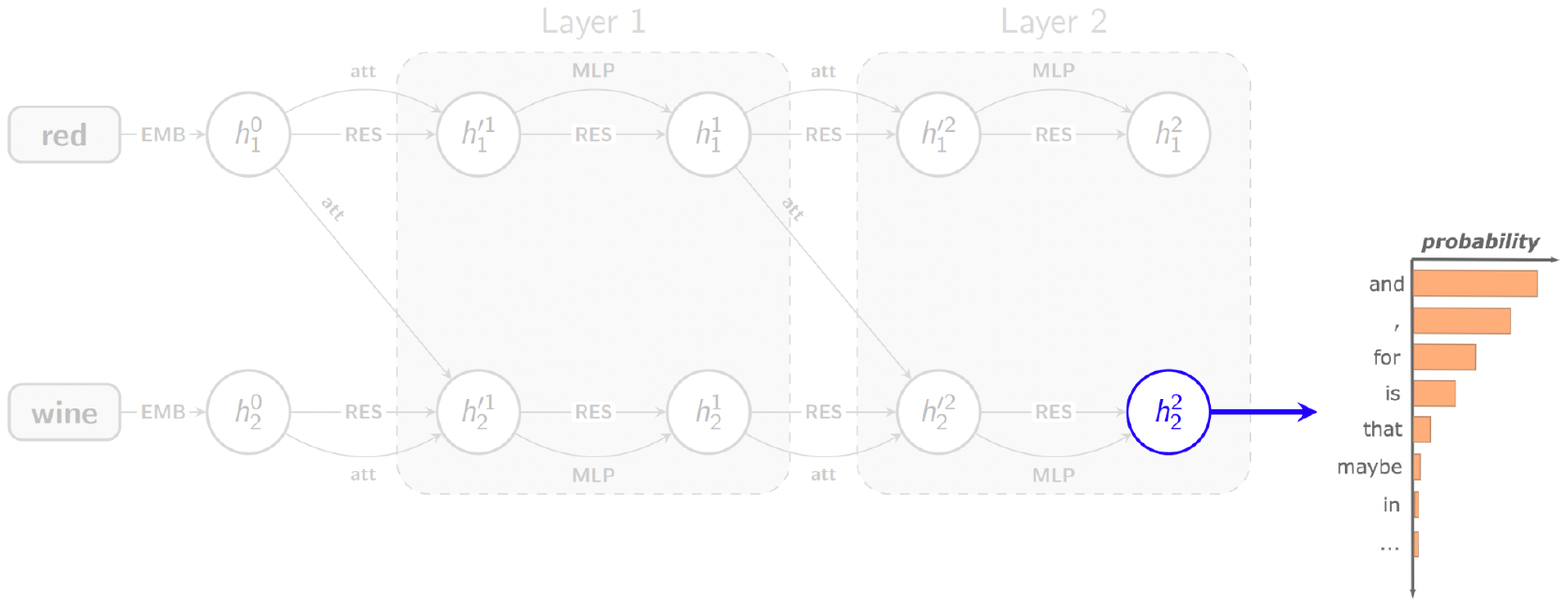}
        \caption{Graph reconstruction starts from the last node.}
        \label{fig:step2}
    \end{subfigure}

    \vspace{1.0em}

    % Row 2
    \begin{subfigure}[t]{0.48\textwidth}
        \centering
        \includegraphics[width=\linewidth]{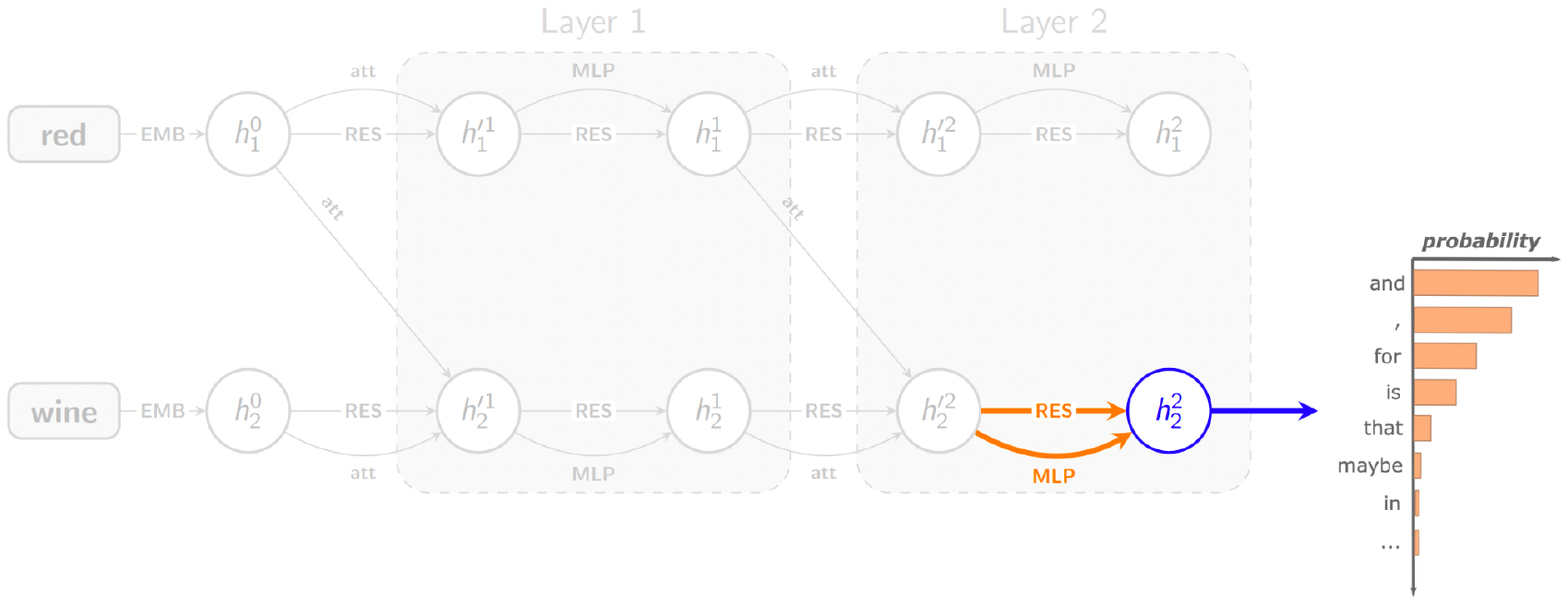}
        \caption{The most important edge is added to the graph.}
        \label{fig:step3_1}
    \end{subfigure}\hfill
    \begin{subfigure}[t]{0.48\textwidth}
        \centering
        \includegraphics[width=\linewidth]{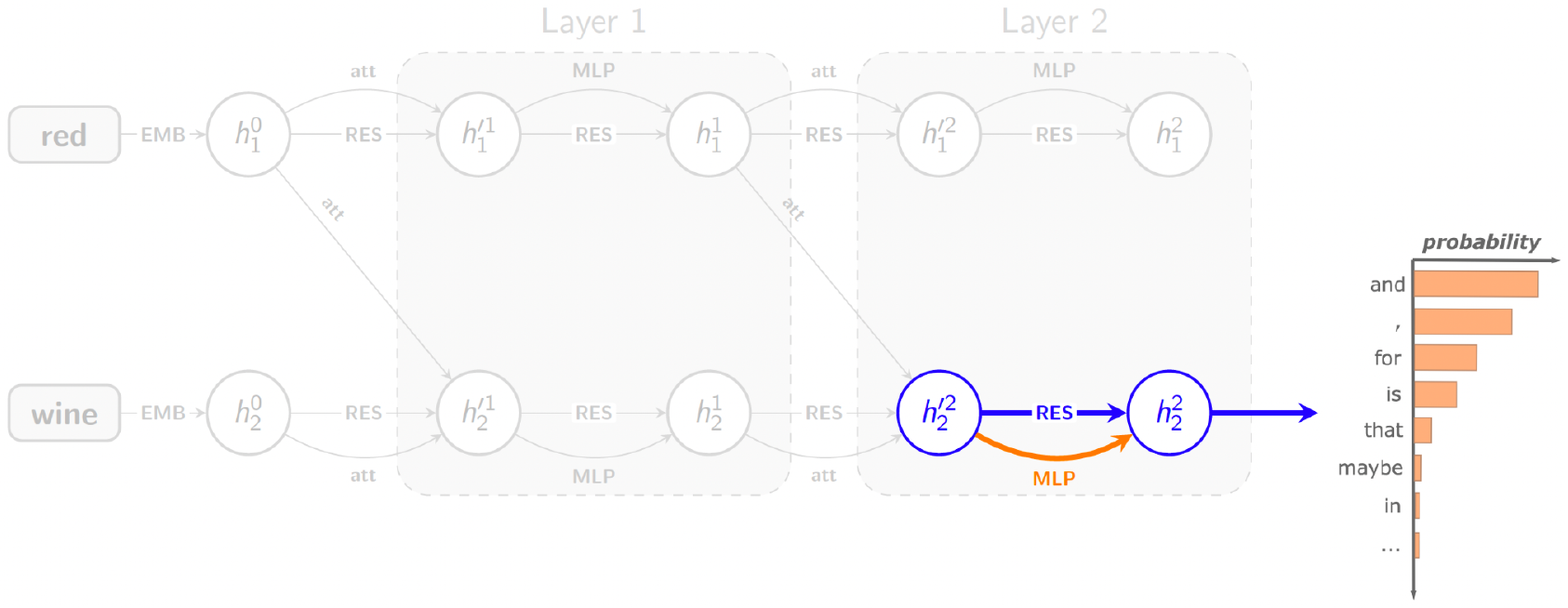}
        \caption{The most important edge is added to the graph.}
        \label{fig:step3_2}
    \end{subfigure}

    \caption{{Graph construction pipeline (Part 1).}}
    \label{fig:graph_construction_part1}
\end{figure*}

\begin{figure*}[t]
    \centering
    % Row 3
    \begin{subfigure}[t]{0.48\textwidth}
        \centering
        \includegraphics[width=\linewidth]{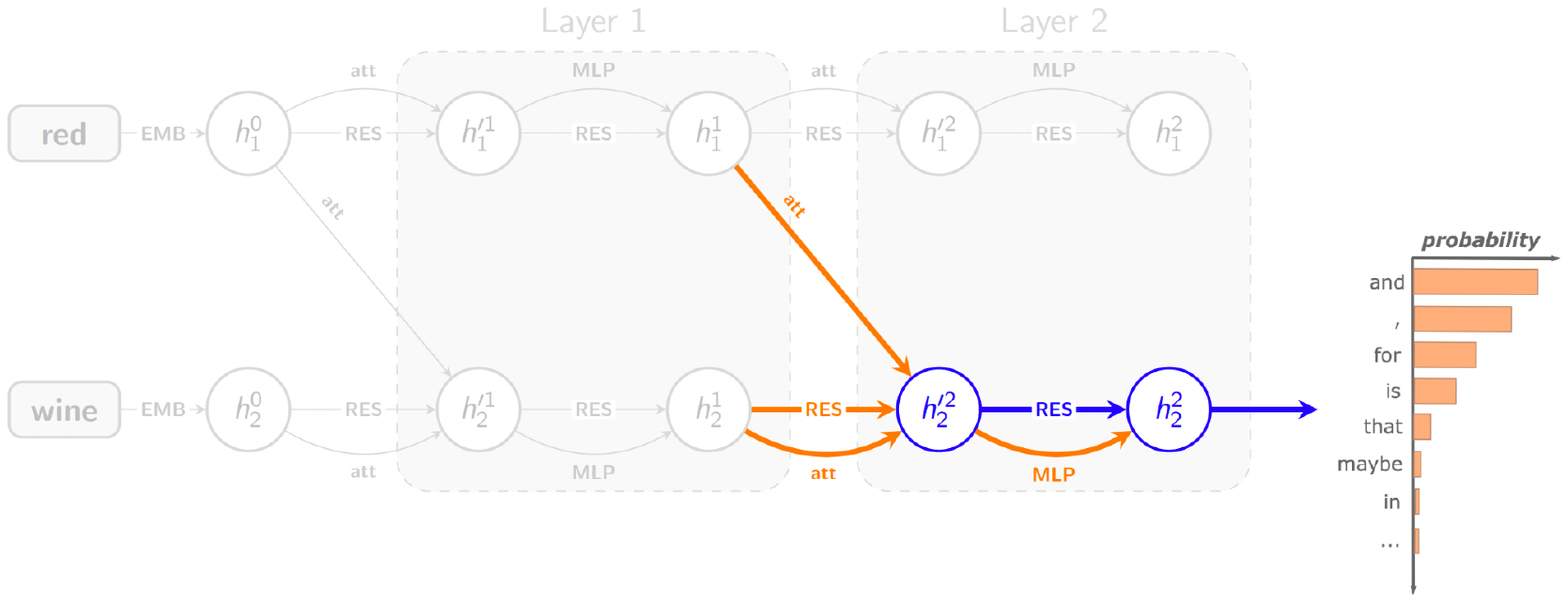}
        \caption{After getting access to the upstream node, the process is repeated.}
        \label{fig:step4}
    \end{subfigure}\hfill
    \begin{subfigure}[t]{0.48\textwidth}
        \centering
        \includegraphics[width=\linewidth]{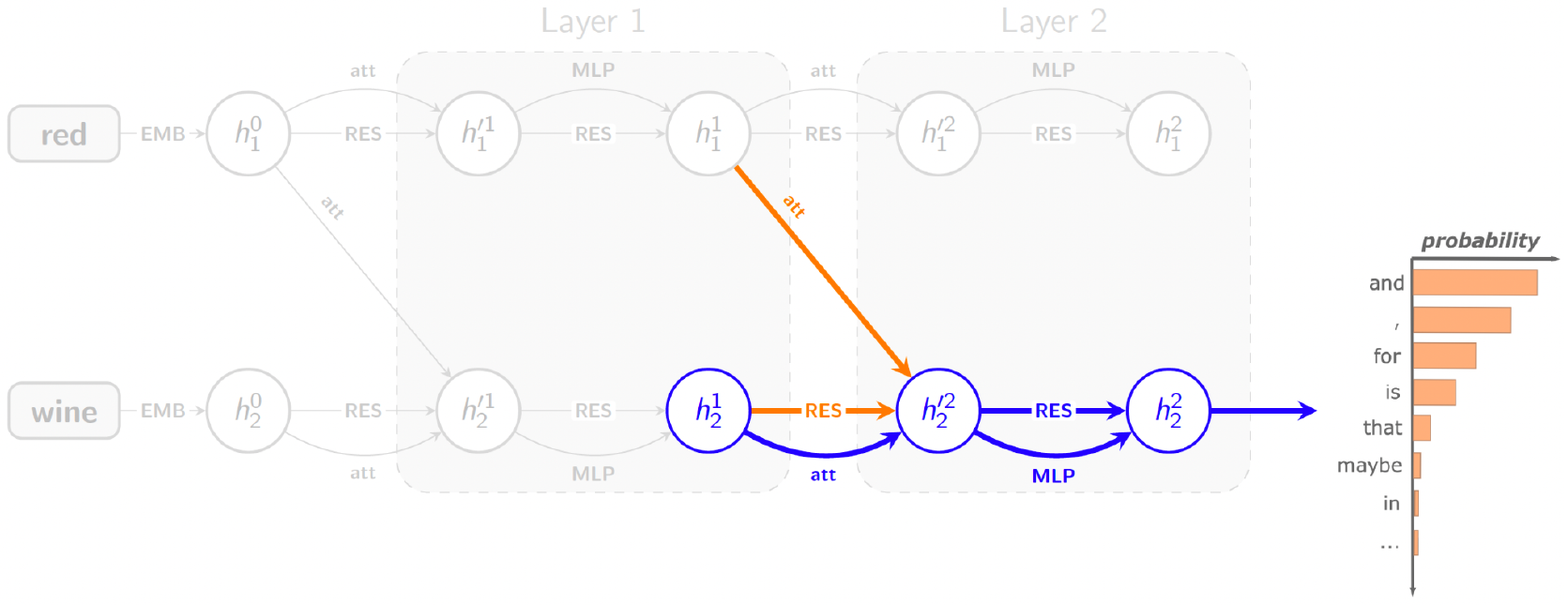}
        \caption{Only the most important edges are being picked and added to the graph until a set 'budget' of possible edges is hit.}
        \label{fig:step5_1}
    \end{subfigure}

    \vspace{1.0em}

    % Row 4
    \begin{subfigure}[t]{0.48\textwidth}
        \centering
        \includegraphics[width=\linewidth]{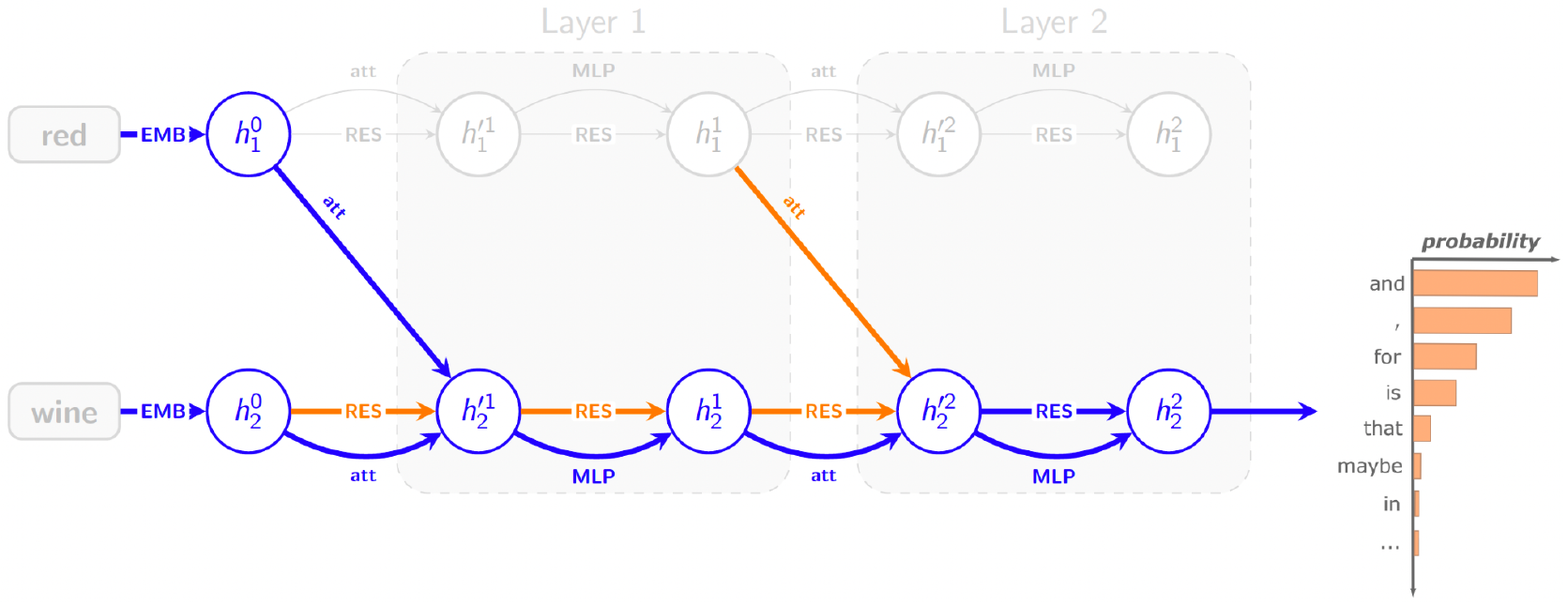}
        \caption{Only the most important edges are being picked and added to the graph until a set 'budget' of possible edges is hit.}
        \label{fig:step5_2}
    \end{subfigure}\hfill
    \begin{subfigure}[t]{0.48\textwidth}
        \centering
        \includegraphics[width=\linewidth]{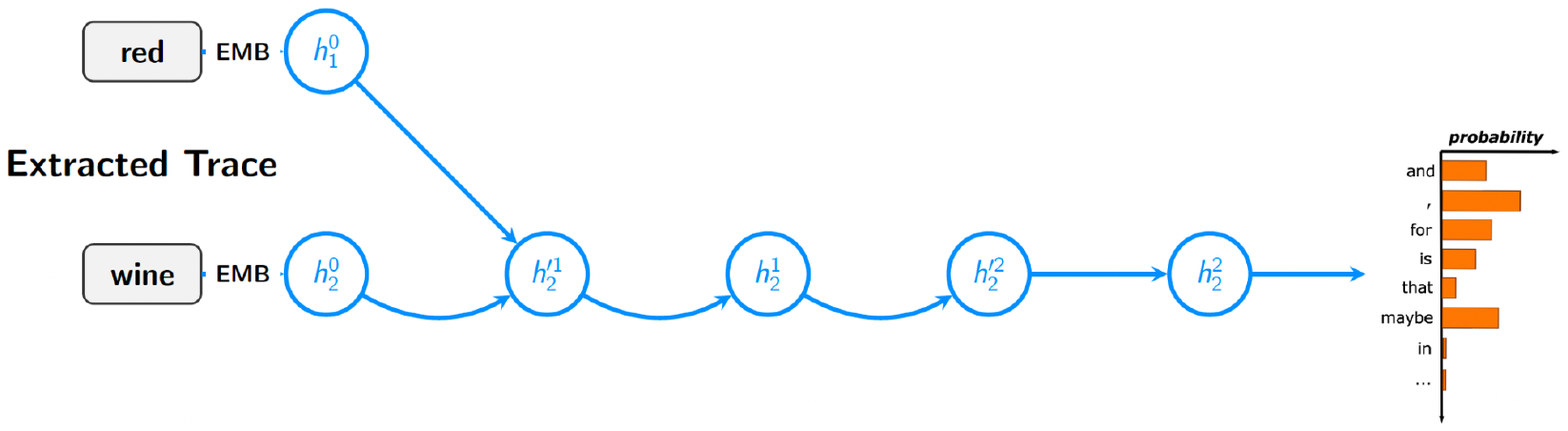}
        \caption{We then compute the reconstruction error by performing a forward pass while masking the edges that do no belong to the graph.}
        \label{fig:step6}
    \end{subfigure}

    \caption{{Graph construction pipeline (Part 2).}}
    \label{fig:graph_construction_part2}
\end{figure*}

The graph traversal algorithm used for trace extraction is described in Algorithm~\ref{alg:best_first_trace} and illustrated in Fig.~\ref{fig:graph_construction_part1} and \ref{fig:graph_construction_part2}.

\begin{algorithm*}
\caption{Greedy BeFS for Trace Construction}
\label{alg:best_first_trace}
\begin{algorithmic}[1]
\Require Computational graph $\mathcal{G} = (\mathcal{V}, \mathcal{E})$, Output node $h^L_n$, Edge budget $s$, Importance function $\mathcal{I}$
\Ensure Trace graph $\mathcal{T}_s = (\mathcal{V}_s, \mathcal{E}_s)$
\State $\mathcal{V}_s \gets \{h^L_n\}$, $\mathcal{E}_s \gets \emptyset$
\State $PQ \gets \text{empty max-priority queue}$
\For{each incoming edge $e = (u, h^L_n) \in \mathcal{E}$ connected to $h^L_n$}
    \State $PQ.\text{insert}(e, \mathcal{I}(e))$ \Comment{Seed the priority queue with initial edges}
\EndFor
\While{$PQ$ is not empty \textbf{and} $|\mathcal{E}_s| < s$}
    \State $e = (u, v) \gets PQ.\text{extract\_max}()$ \Comment{Select edge with highest score}
    \State $\mathcal{E}_s \gets \mathcal{E}_s \cup \{e\}$
    \If{$u \notin \mathcal{V}_s$}
        \State $\mathcal{V}_s \gets \mathcal{V}_s \cup \{u\}$
        \For{each incoming edge $e' = (w, u) \in \mathcal{E}$ connected to $u$}
            \State $PQ.\text{insert}(e', \mathcal{I}(e'))$ \Comment{Expand frontier}
        \EndFor
    \EndIf
\EndWhile
\State \Return $\mathcal{T}_s = (\mathcal{V}_s, \mathcal{E}_s)$
\end{algorithmic}
\end{algorithm*}

\section{Code and Resources}
We developed the code for extracting and evaluating the trace from Hugging Face\footnote{\url{https://huggingface.co/}} models. We modified LLM's modeling source code from Hugging Face in order to (1) extract LLM's internal representations when constructing the computational graph, and (2) ablate edges during the evaluation. The code and data required to reproduce the experiments will be made publicly available on a Github repository upon acceptance.

\section{Computation Profile}

We provide benchmarking numbers for an OLMo-13B model with 40-word instances executed on an NVIDIA A30 GPU:
\begin{itemize}
    \item \textbf{Step 1} (GPU - Importance Scoring of the Edges): 100 instances take 2h 10m (one single forward pass per instance to collect magnitude scores).
    \item \textbf{Step 2} (CPU - Greedy Trace Traversal): Extracts the traces across all 26 size budgets for 100 instances in 50m.
    \item \textbf{Step 3} (GPU - Faithfulness Evaluation): Performs 26 evaluation forward passes per instance to measure the TV curves, taking 1d 8h per 100 instances.
\end{itemize}
        
For our full corpus of 5,000 instances, Step 1 requires 4d12h20m, Step 2 1d17h40m, and Step 3 66d6h. Trace extraction (Steps 1 \& 2) is rather fast; most of the cost comes from evaluating faithfulness.

\section{Methods: Additional Information}
\label{app:method}

\subsection{Estimating Computational Density}
\label{app:rho}
To capture the global sparsity of a model across different relative trace sizes $s$, we define the scalar \textit{Computational Density} metric $\mathcal{C}$ as the Area Under the Curve of the reconstruction error plotted against the log of the trace size. This metric is \textit{instance-wise}, meaning that we get one $\mathcal{C}$ per model prediction in the dataset. We obtain it by evaluating the traces on a fixed grid of size $S = \{s_0, s_1, \dots, s_K\}$, where $0 = s_0 < s_1 < \dots < s_K = 1$. We estimate $\mathcal{C}$ using the trapezoidal rule:
\begin{equation}
\label{eq:trapezoidal}
    D = \frac{1}{2} \sum_{k=1}^{K} (s_k - s_{k-1}) \cdot [\epsilon(s_k) + \epsilon(s_{k-1})]
\end{equation}
where $\epsilon(s) = \delta_{TV}(s)$ is the reconstruction error for a trace of size $s$. Intuitively, $\mathcal{C}$ measures the model's resistance to pruning:\\
\textit{Low $\mathcal{C}$ (Sparse):} The error $\epsilon(s)$ drops rapidly to zero even for small $s$, implying that a few edges suffice to replicate the computation;\\
\textit{High $\mathcal{C}$ (Dense):} The error remains significant until $s$ approaches 1, implying that information is distributed diffusely across the graph.

\subsection{Grid Size For Trace Extraction}
\label{app:grid-size}
In order to extract the trace at different sizes, we empirically set 
$s \in [
1e^{-5}, 2e^{-5}, 4e^{-5}, 8e^{-5}, 1e^{-4}, 2e^{-4}, 4e^{-4}, 8e^{-4},$
$1e^{-3}, 1.2e^{-3}, 1.4e^{-3}, 2e^{-3}, 3e^{-3}, 4e^{-3}, 6e^{-3}, $
$8e^{-3}, 1e^{-2}, 2e^{-2}, 4e^{-2}, 6e^{-2}, 8e^{-2}, 1e^{-1}, 2e^{-1},$ $4e^{-1}, 6e^{-1}, 8e^{-1}]$. These values span diverse trace sizes from sparse to dense, and reconstruction errors, from random to almost-perfect reconstruction. They were obtained by manually testing them on a few instances.

\subsection{Model Zoo}
\label{app:model-zoo}
The LLMs used in this work are detailed in Table~\ref{tab:model_summary}.

\begin{table*}[t] % The '*' makes the table span both columns cleanly
\centering
\small 
\caption{Overview of the models used in this study. All models are decoder-only Transformers. \textit{Parameters} refers to total parameters. Training data size is denoted in tokens ($T$ for Trillion, $B$ for Billion).}
\label{tab:model_summary}
\begin{tabularx}{\textwidth}{l c c >{\raggedright\arraybackslash}X r}
\toprule
\textbf{Model Name} & \textbf{Parameters} & \textbf{Tokens} & \textbf{Hugging Face ID} & \textbf{\# (Valid Instances)}\\
\midrule

OLMo-2-1124-7B~\cite{olmo20242} & 7.3B & 4.0T & \texttt{allenai/OLMo-2-1124-7B}& 4921\\
OLMo-2-1124-13B~\cite{olmo20242} & 13.7B & 5.0T & \texttt{allenai/OLMo-2-1124-13B}& 4975\\
\addlinespace[0.3em]
Qwen3-8B-Base~\cite{yang2025qwen3} & 8.2B & 36.0T & \texttt{Qwen/Qwen3-8B-Base}& 4944\\
Qwen3-14B-Base~\cite{yang2025qwen3} & 14.8B & 36.0T & \texttt{Qwen/Qwen3-14B-Base}& 4959\\
\addlinespace[0.3em]
Qwen2.5-7B~\cite{yang2025qwen3} & 7.6B & 18.0T & \texttt{Qwen/Qwen2.5-7B}& 4954\\
Qwen2.5-14B~\cite{yang2025qwen3} & 14.7B & 18.0T & \texttt{Qwen/Qwen2.5-14B} & 4275\\
\addlinespace[0.3em]
Llama-3.1-8B~\cite{grattafiori2024llama} & 8.0B & 15.0T & \texttt{meta-llama/Llama-3.1-8B}& 4953\\
\addlinespace[0.3em]
Mistral-7B-v0.1~\cite{jiang2024mistral} & 7.3B & Not disclosed & \texttt{mistralai/Mistral-7B-v0.1}& 5000\\
\addlinespace[0.3em]
DeepSeek-LLM-7B~\cite{bi2024deepseek} & 7.0B & 2.0T & \texttt{deepseek-ai/deepseek-llm-7b-base}& 4893\\
\addlinespace[0.3em]
Phi-4~\cite{abdin2024phi} & 14.0B & 9.8T & \texttt{microsoft/phi-4}& 4964\\
\bottomrule
\end{tabularx}
\end{table*}

\section{Additional Results}

\subsection{Formal Definition of the Phases}
\label{app:formal_phases}
We provide a formal definition of the phase transitions based on the normalized derivative of the Total Variation ($TV$) distance curve with respect to the trace size $s$. We look at the regime where the $TV$ distance decreases rapidly as the trace size expands, satisfying the following condition: $$\frac{dTV/ds}{\max (dTV/ds)} < -0.2$$
The Construction Phase corresponds to the interval during which this condition is met. Averaging across models yields stable boundaries: the construction phase starts at $s = 2.15 \times 10^{-4}$ and stops at $s = 1.67 \times 10^{-2}$. Below this final threshold, the model enters the Refinement Phase.

\subsection{Random trace extraction baseline}
\label{app:rand_baseline}
\begin{figure}
    \centering
    \includegraphics[width=0.9\linewidth]{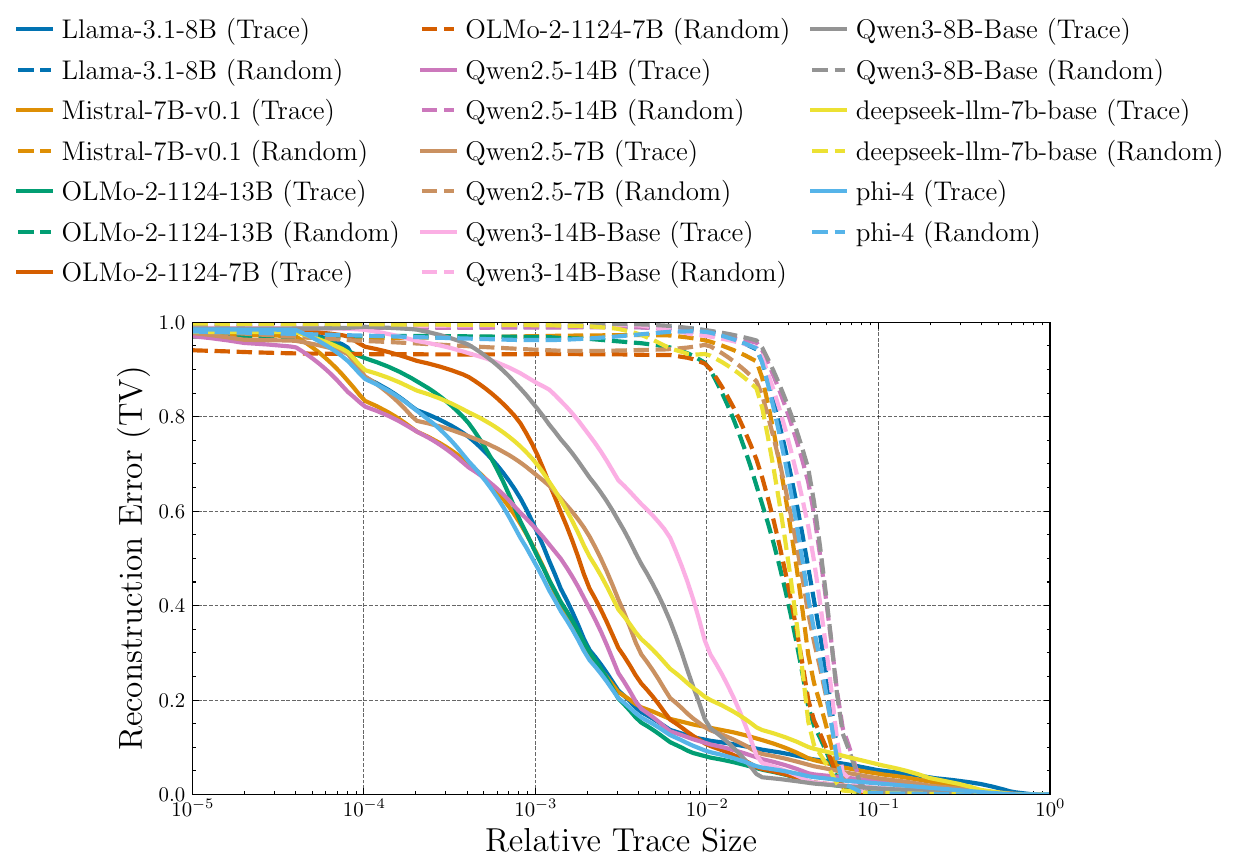}
    \caption{We compare our trace extraction with a strong random baseline that selects the MLPs and residual edges first. The trace extraction method used in this paper outperforms the random baseline at any size.}
    \label{fig:rand}
\end{figure}

As a sanity check, we compare our trace extraction method to a random one. The approach is identical to Algorithm~\ref{alg:best_first_trace}, but the importance function is replaced by a random attribution that samples importance weights from a uniform distribution. 
As the proportion of residual, MLP and attention edges is highly unbalanced (dominated by attention), a naive random sampling would disfavor residuals and MLPs. Therefore, we use a stronger baseline that, at each step, selects the residual and MLPs edges first. We use the same inputs described in Section~\ref{sec:methodology}. We observe that the trace extraction method used in this paper outperforms the random baseline at any size (for an equivalent reconstruction score, our method leads to a smaller trace in average). Results are shown in Fig.~\ref{fig:rand}. 

\subsection{Sufficient vs.~Necessary}
\label{app:suf_nec}
\begin{figure}
    \centering
    \includegraphics[width=0.98\linewidth]{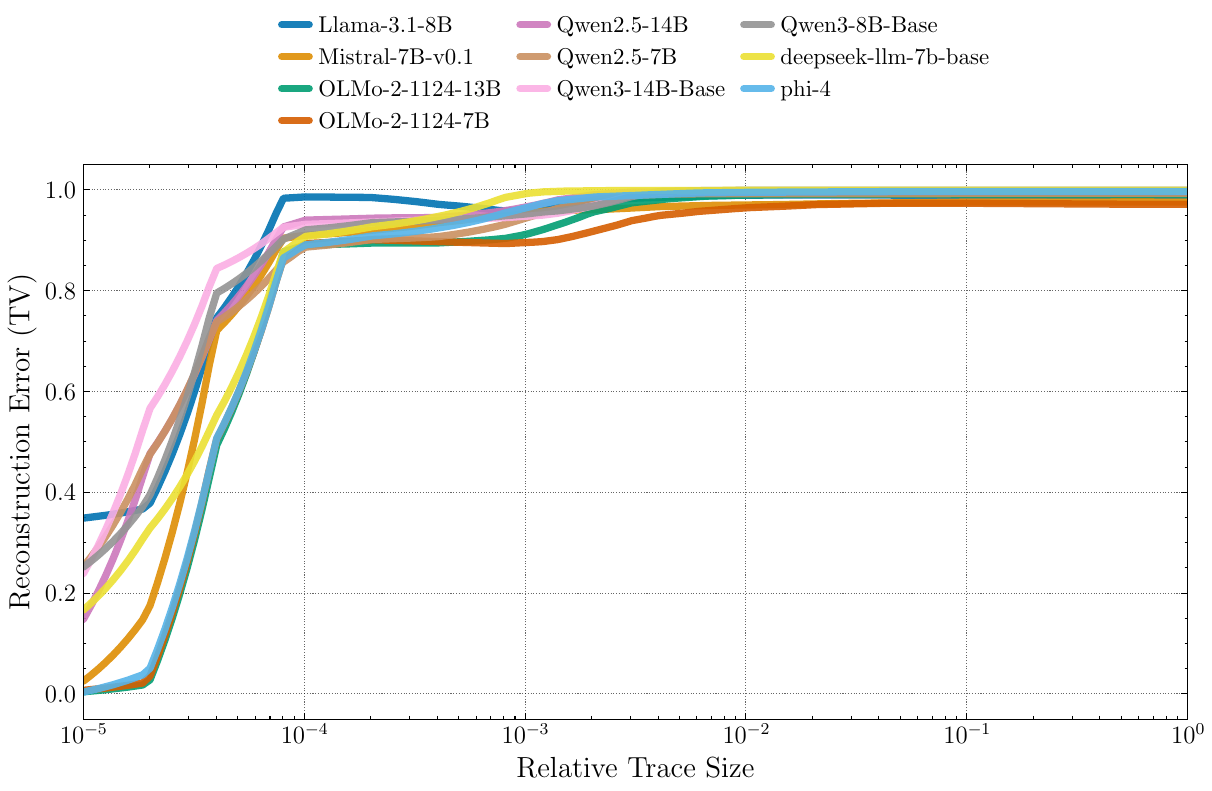}
    \caption{We measure the effect of zeroing the edges belonging to the trace. At $s<10^{-4}$, the reconstruction error is low, demonstrating the resilience of LLMs to a small number of edge ablations. However, quickly (starting from $s=10^{-4}$), ablating the trace leads to a very high reconstruction error, indicating that these edges are necessary to reconstruct the prediction. This contrasts with Fig.~\ref{fig:tv_size}, which illustrates the complementary effect of zeroing edges that do \textit{not} belong to the trace.}
    \label{fig:tv_inverse}
\end{figure}

As a second sanity check, we measure the effect of ablating only the trace. Specifically, we zero the edges in the computational graph that belong to the trace (instead of keeping only those edges). Because it is trivial that ablating residual edges would lead to a strong drop in performance, we never ablate them in this setup. Thus, we measure the effect of ablating only the MLP and attention edges that are in the trace. We use the same inputs described in Section~\ref{sec:methodology}. Results are shown in Fig.~\ref{fig:tv_inverse}. We observe that the edges identified by our trace extraction method are indeed necessary to reconstruct a faithful output.

\subsection{Density vs.~Entropy, Loss and Frequency}
\label{app:entropy}

\begin{figure}
    \centering
    \includegraphics[width=0.98\linewidth]{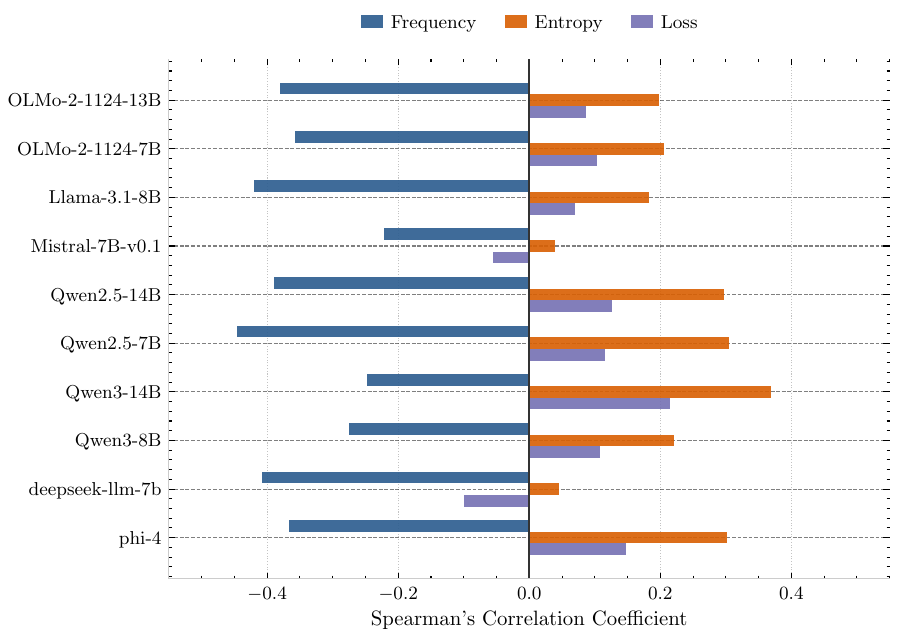}
    \caption{Spearman correlation between computation density and: the entropy of the output probability distribution, the language modeling loss, and the frequency of the top-1 predicted token. All correlations are highly statistically significant ($p < 0.01$).}
    \label{fig:density_ent}
\end{figure}

Fig.~\ref{fig:density_ent} shows the correlation between density and the entropy of the LLM output, the Wikitext-based frequency of the input, and the loss.

\subsection{Pairwise correlation of LLMs}
\label{app:pairwise_corr}

\begin{figure}[H] % Note the capital H
    \centering
    \includegraphics[width=0.98\linewidth]{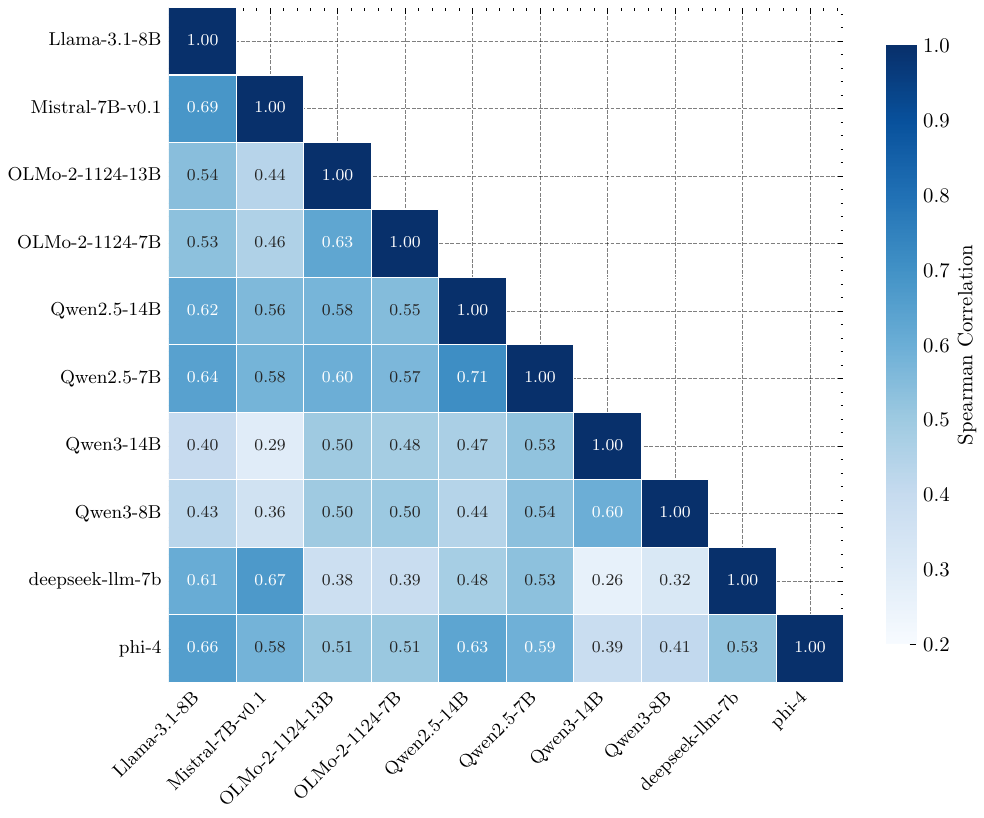}
    \caption{We compute the Spearman correlation of $\mathcal{C}$ across pairs of distinct language models evaluated on identical sentences.}
    \label{fig:pairwise_corr}
\end{figure}

Fig.~\ref{fig:pairwise_corr} shows the Spearman correlation of $\mathcal{C}$ across pairs of distinct language models evaluated on identical sentences.

\subsection{Alternative Importance Scoring Functions}
\label{app:alternative_scores}

In the main text, we extract the computational trace using a node-normalized L1-norm to compute the importance score $\mathcal{I}(e)$. To validate this choice, we experimented with several alternative scoring functions, assessing their ability to yield low reconstruction error at comparable trace sizes (cf. Fig.).

\begin{figure*}
    \centering
    \includegraphics[width=\linewidth]{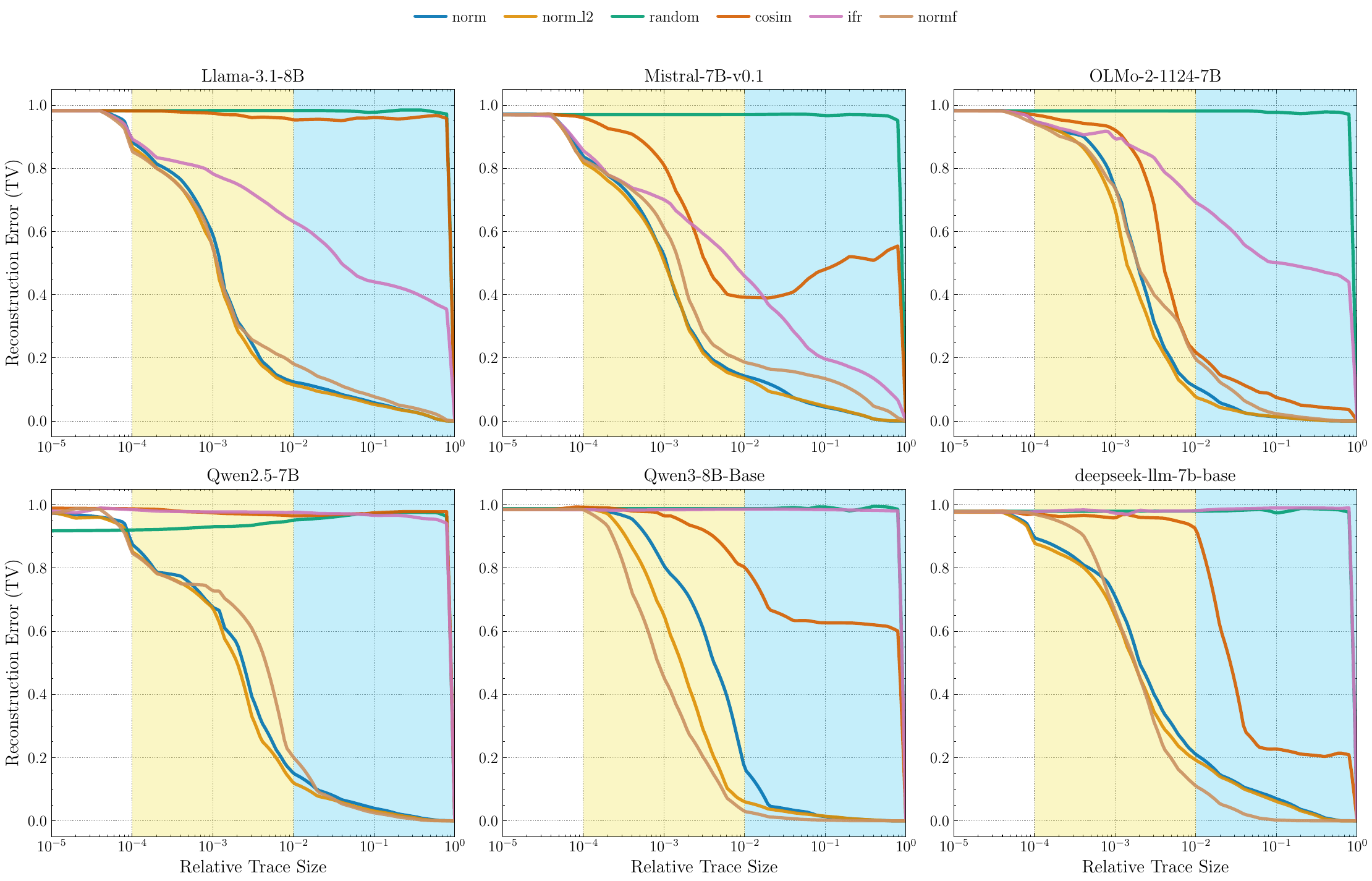}
    \caption{Alternative importance scoring functions (see description of the scoring functions in Appendix~\ref{app:alternative_scores}).}
    \label{fig:alternative_scores}
\end{figure*}

For the formal definitions below, let $v$ be a target node in the computational graph with representation $\mathbf{v}_v$. Let $e$ be an incoming edge to $v$ carrying the update vector $\mathbf{v}_e$, and let $\text{in}(v)$ denote the set of all incoming edges to $v$.

\paragraph{L2-Norm (\texttt{norm\_l2})} 
Similar to our primary metric, this approach assumes the importance of an operation is proportional to its magnitude, but penalizes smaller activations more heavily by using the Euclidean norm:
$$ \mathcal{I}_{\text{L2}}(e) = \frac{\| \mathbf{v}_e \|_2}{\sum_{e' \in \text{in}(v)} \| \mathbf{v}_{e'} \|_2} $$

\paragraph{Target-Normalized Norm (\texttt{normf})} 
Instead of normalizing an edge's magnitude by the sum of all competing updates, this variant normalizes the edge directly by the magnitude of the final target node representation:
$$ \mathcal{I}_{\text{target-norm}}(e) = \frac{\| \mathbf{v}_e \|_1}{\| \mathbf{v}_v \|_1} $$

\paragraph{Proximity Score (\texttt{ifr})} 
This metric, based on IFR~\cite{ferrando2024information}, measures how much a specific edge vector moves the representation toward the final target node vector. We compute the distance between the edge and the target, subtracted from the target's baseline norm. To ensure all scores are positive before node-wise normalization, we shift the distribution by its minimum value:
$$ d(e) = \| \mathbf{v}_v \|_p - \| \mathbf{v}_e - \mathbf{v}_v \|_p $$
$$ \mathcal{I}_{\text{prox}}(e) = \frac{d(e) + |\min_{e'} d(e')|}{\sum_{e' \in \text{in}(v)} \left( d(e') + |\min_{e''} d(e'')| \right)} $$

\paragraph{Cosine Similarity (\texttt{cosim})} 
This approach ignores the absolute magnitude of the update, focusing solely on its angular alignment with the final target node representation. We map the standard cosine similarity space $[-1, 1]$ to $[0, 1]$ so that orthogonal updates receive a score of $0.5$ and perfectly aligned updates receive $1$:
$$ \mathcal{I}_{\text{cos}}(e) = \frac{1}{2} \left( 1 + \frac{\mathbf{v}_e \cdot \mathbf{v}_v}{\| \mathbf{v}_e \|_2 \| \mathbf{v}_v \|_2} \right) $$

% \paragraph{Leverage Scores (\texttt{lev})} 
% Drawing on matrix approximation techniques, we compute the statistical leverage scores of the incoming edge vectors over rank $k$. This highlights edges that contribute unique, high-variance directions to the target node's subspace. The scores are then normalized over $\text{in}(v)$ to sum to $1$.

\paragraph{Random Baseline (\texttt{random})} 
To establish a baseline for trace extraction, we completely decouple the score from the computational properties of the graph, assigning uniformly random importance:
$$ \mathcal{I}_{\text{rand}}(e) \sim \mathcal{U}(0, 1) $$

\vspace{1em}
\noindent \textbf{Summary of Results.} We found that magnitude-based metrics generally outperformed alignment-based metrics (like cosine similarity), suggesting that the scale of an update is a stronger proxy for its necessity than its direction.

\subsection{Structure of the Trace (all models)}
\label{app:structure}
Fig~\ref{fig:all_pages_comparison} and \ref{fig:all_pages_comparison_2} provide the structure of the trace for all models tested in this study.
\subsection{Component frequency (all models)}
\label{app:comp_freq}
Fig~\ref{fig:all_frequency_comparison} and \ref{fig:all_frequency_comparison_2} provide component frequency at different sizes for all models.

\clearpage

\onecolumn
% --- Start of multi-page aligned loop ---
\begin{figure*}
    \begin{center}
    % The \foreach loop: change "10" to the actual total number of pages in your PDF
    \foreach \p in {1,...,5} {
        
        % Removing \begin{figure} makes these elements static instead of floating
        \begin{minipage}{\textwidth}
            \centering
            
            % Left PDF
            \begin{subfigure}{0.48\textwidth}
                \centering
                \includegraphics[width=\linewidth, page=\p]{figures-emnlp-draft/fig4_layer_composition_with_nu_40.pdf}
            \end{subfigure}%
            \hfill% % Space between images
            % Right PDF
            \begin{subfigure}{0.48\textwidth}
                \centering
                \includegraphics[width=\linewidth, page=\p]{figures-emnlp-draft/fig5_component_composition_with_nu_40.pdf}
            \end{subfigure}
            
            %\vspace{-10pt} % Adds vertical space between the rows
        \end{minipage}\par % Allows LaTeX to break to a new page between rows if needed
        % Removing \end{figure}
    }
    
    % Main caption and label for the entire multi-page collection
    \captionof{figure}{Structure of the trace at different sizes.}
    \label{fig:all_pages_comparison}
    \end{center}
\end{figure*}

\begin{figure*}
    \begin{center}
    % The \foreach loop: change "10" to the actual total number of pages in your PDF
    \foreach \p in {6,...,10} {
        
        % Removing \begin{figure} makes these elements static instead of floating
        \begin{minipage}{\textwidth}
            \centering
            
            % Left PDF
            \begin{subfigure}{0.48\textwidth}
                \centering
                \includegraphics[width=\linewidth, page=\p]{figures-emnlp-draft/fig4_layer_composition_with_nu_40.pdf}
            \end{subfigure}%
            \hfill% % Space between images
            % Right PDF
            \begin{subfigure}{0.48\textwidth}
                \centering
                \includegraphics[width=\linewidth, page=\p]{figures-emnlp-draft/fig5_component_composition_with_nu_40.pdf}
            \end{subfigure}
            
            %\vspace{-10pt} % Adds vertical space between the rows
        \end{minipage}\par % Allows LaTeX to break to a new page between rows if needed
        % Removing \end{figure}
    }
    
    % Main caption and label for the entire multi-page collection
    \captionof{figure}{Structure of the trace at different sizes.}
    \label{fig:all_pages_comparison_2}
    \end{center}
\end{figure*}

\clearpage

% --- Start of multi-page aligned loop ---

\begin{figure*}
    \begin{center}
    % Loop through odd numbers up to 9. 
    % \nextp is automatically evaluated as \p + 1.
    \foreach \p [evaluate=\p as \nextp using int(\p+1)] in {1,3} {
        
        % Minipage keeps these two side-by-side images on the same row
        \begin{minipage}{\textwidth}
            \centering
            
            % Left PDF (Odd page)
            \begin{subfigure}{0.41\textwidth}
                \centering
                \includegraphics[width=\linewidth, page=\p]{figures-emnlp-draft/fig6_component_frequency_40.pdf}
                % \caption{Page \p}
                \label{fig:page_\p}
            \end{subfigure}% <-- Note the % here to prevent a space
            \hfill% <-- Note the % here to prevent a space
            % Right PDF (Even page)
            \begin{subfigure}{0.41\textwidth}
                \centering
                \includegraphics[width=\linewidth, page=\nextp]{figures-emnlp-draft/fig6_component_frequency_40.pdf}
                % \caption{Page \nextp}
                \label{fig:page_\nextp}
            \end{subfigure}
            
            \vspace{-15pt} % Adds vertical space between the rows
        \end{minipage}\par % Allows LaTeX to break to a new page between rows if needed
    }
    
    % Main caption and label for the entire multi-page collection
    \captionof{figure}{Component frequency at different sizes for all models.}
    \label{fig:all_frequency_comparison}
    \end{center}
\end{figure*}

\begin{figure*}
    \begin{center}
    % Loop through odd numbers up to 9. 
    % \nextp is automatically evaluated as \p + 1.
    \foreach \p [evaluate=\p as \nextp using int(\p+1)] in {6,9} {
        
        % Minipage keeps these two side-by-side images on the same row
        \begin{minipage}{\textwidth}
            \centering
            
            % Left PDF (Odd page)
            \begin{subfigure}{0.41\textwidth}
                \centering
                \includegraphics[width=\linewidth, page=\p]{figures-emnlp-draft/fig6_component_frequency_40.pdf}
                % \caption{Page \p}
                \label{fig:page_\p}
            \end{subfigure}% <-- Note the % here to prevent a space
            \hfill% <-- Note the % here to prevent a space
            % Right PDF (Even page)
            \begin{subfigure}{0.41\textwidth}
                \centering
                \includegraphics[width=\linewidth, page=\nextp]{figures-emnlp-draft/fig6_component_frequency_40.pdf}
                % \caption{Page \nextp}
                \label{fig:page_\nextp}
            \end{subfigure}
            
            \vspace{-15pt} % Adds vertical space between the rows
        \end{minipage}\par % Allows LaTeX to break to a new page between rows if needed
    }
    
    % Main caption and label for the entire multi-page collection
    \captionof{figure}{Component frequency at different sizes for all models.}
    \label{fig:all_frequency_comparison_2}
    \end{center}
\end{figure*}

% --- End of multi-page aligned loop ---

% ==========================================
% FIGURE 1: PART 1 (Steps 1 to 3.2 - 2x2 grid)
% ==========================================

\clearpage % Forces the second 2x2 grid to top of next page

% ==========================================
% FIGURE 2: PART 2 (Steps 4 to 6 - 2x2 grid)
% ==========================================

\end{document}